\documentclass[fleqn,10pt]{wlscirep}
\usepackage[utf8]{inputenc}
\usepackage[T1]{fontenc}
\usepackage{multirow}
\usepackage{amsmath, amsfonts, amssymb, amsthm}
\usepackage{newtxmath} 

\title{IXGS---Intraoperative 3D Reconstruction from Sparse, Arbitrarily Posed Real X-rays}

\author[1,*]{Sascha Jecklin}
\author[1]{Aidana Massalimova}
\author[3]{Ruyi Zha}
\author[1]{Lilian Calvet}
\author[2]{Christoph J. Laux}
\author[2]{Mazda Farshad}
\author[1]{Philipp Fürnstahl}
\affil[1]{Research in Orthopedic Computer Science, Balgrist University Hospital, Zurich, 8008, Switzerland}
\affil[2]{Department of Orthopedics, Balgrist University Hospital, University of Zurich, Zurich, 8008, Switzerland}
\affil[3]{The Australian National University, Canberra, ACT 2601, Australia}

\affil[*]{sascha.jecklin@balgrist.ch}


    \begin{abstract}
        Spine surgery is a high-risk intervention demanding precise execution,
        often supported by image-based navigation systems. Recently, supervised
        learning approaches have gained attention for reconstructing 3D spinal
        anatomy from sparse fluoroscopic data, significantly reducing reliance on
        radiation-intensive 3D imaging systems. However, these methods typically
        require large amounts of annotated training data and may struggle to generalize
        across varying patient anatomies or imaging conditions. Instance-learning
        approaches like Gaussian splatting could offer an alternative by
        avoiding extensive annotation requirements. While Gaussian splatting has
        shown promise for novel view synthesis, its
        application to sparse, arbitrarily posed real intraoperative X-rays has remained
        largely unexplored.

        This work addresses this limitation by extending the $R^{2}$-Gaussian splatting
        framework to reconstruct anatomically consistent 3D volumes under
        these challenging conditions. We introduce an anatomy-guided
        radiographic standardization step using style transfer, improving visual
        consistency across views, and enhancing reconstruction quality. Notably, our
        framework requires no pretraining, making it inherently adaptable to new
        patients and anatomies.

        We evaluated our approach using an \textit{ex-vivo} dataset. Expert surgical evaluation
        confirmed the clinical utility of the 3D reconstructions for navigation,
        especially when using 20 to 30 views, and highlighted the
        standardization's benefit for anatomical clarity. Benchmarking via quantitative
        2D metrics (PSNR/SSIM) confirmed performance trade-offs compared to idealized
        settings, but also validated the improvement gained from standardization over
        raw inputs.

        This work demonstrates the feasibility of instance-based volumetric reconstruction
        from arbitrary sparse-view X-rays, advancing intraoperative 3D imaging
        for surgical navigation. Code and data to reproduce our results is made
        available at \url{https://github.com/MrMonk3y/IXGS}.
    \end{abstract}
\begin{document}

\flushbottom
\maketitle

\section*{Introduction}
    Studies have demonstrated that surgical navigation can contribute to an improved
    patient outcome, reduce complications, shorten operating time, and lower
    revision rates \cite{tonetti_role_2020,keil_intraoperative_2023}. Clinically
    established navigation systems primarily rely on either preoperative imaging
    registered to the patient's anatomy during surgery or directly utilize intraoperative
    3D imaging for guidance.

    Widely used registration-based approaches utilize point-, contour-, or intensity-based
    registration, where preoperative computed tomography (CT) data is registered
    to the patient's anatomy during surgery
    \cite{hong_effective_2010,ma_autonomous_2020,suenaga_vision-based_2015}. Deep
    learning-based methods have been proposed to improve registration accuracy and
    robustness \cite{zheng_pairwise_2018,ferrante_adaptability_2018}. However, registration
    remains a highly ill-posed problem, making the process susceptible to data
    ambiguity, user-dependent variability, and anatomical changes
    \cite{zhang_risk_2019,rahmathulla_intraoperative_2014}.

    Intraoperative 3D imaging devices such as cone-beam computed tomography (CBCT)
    enable high-resolution intraoperative 3D reconstructions for spinal
    procedures and eliminates the need for registration. However, this advantage
    comes at the cost of increased radiation exposure. CBCT typically requires between
    180 and 1024 projections to reconstruct a 3D volume \cite{venkatesh_cone_2017}.
    The patient radiation dose for spinal procedures using CBCT can increase by
    2.8 to 9.96 times compared to conventional 2D fluoroscopy
    \cite{costa_spinal_2011, mendelsohn_patient_2016, villard_radiation_2014}. Beyond radiation risks,
    CBCT also has technical limitations such as a restricted field of view, more
    elaborate draping requirements, and longer acquisition times compared to
    conventional 2D fluoroscopy
    \cite{dea_economic_2016,costa_spinal_2011,tonetti_role_2020,beck_benefit_2009}.
    Reducing intraoperative radiation while maintaining high-quality 3D reconstructions
    is therefore a key objective.

    Previous research has focused on preserving the benefits of high-quality (CBCT-like)
    intraoperative 3D reconstruction while reducing radiation exposure,
    particularly in spinal procedures. In our prior work, X23D, we demonstrated that
    a neural network can accurately reconstruct a 3D volume of the lumbar spine
    from only four X-rays using a neural network, achieving high accuracy with
    low radiation exposure
    \cite{jecklin_x23dintraoperative_2022,jecklin_domain_2024}. X23D integrates prior
    anatomical knowledge through end-to-end training on a large, curated dataset.
    However, this approach limits its scalability, particularly when accommodating
    the full spectrum of anatomical variations, pathological conditions, and implant
    types encountered in the operating room.

    Neural scene representation techniques overcome these limitations, as they
    are capable of synthesizing high-quality 3D representations from
    sparse 2D data without requiring pretraining \cite{zha_r2-gaussian_2024,zha_naf_2022,ruckert_neat_2022}.
    Compared to our previous work, X23D, their architecture allows for processing
    larger images on similar hardware, accommodating larger scenes without
    losing resolution.
    \begin{figure}
        \centering
        \includegraphics[width=0.7\textwidth]{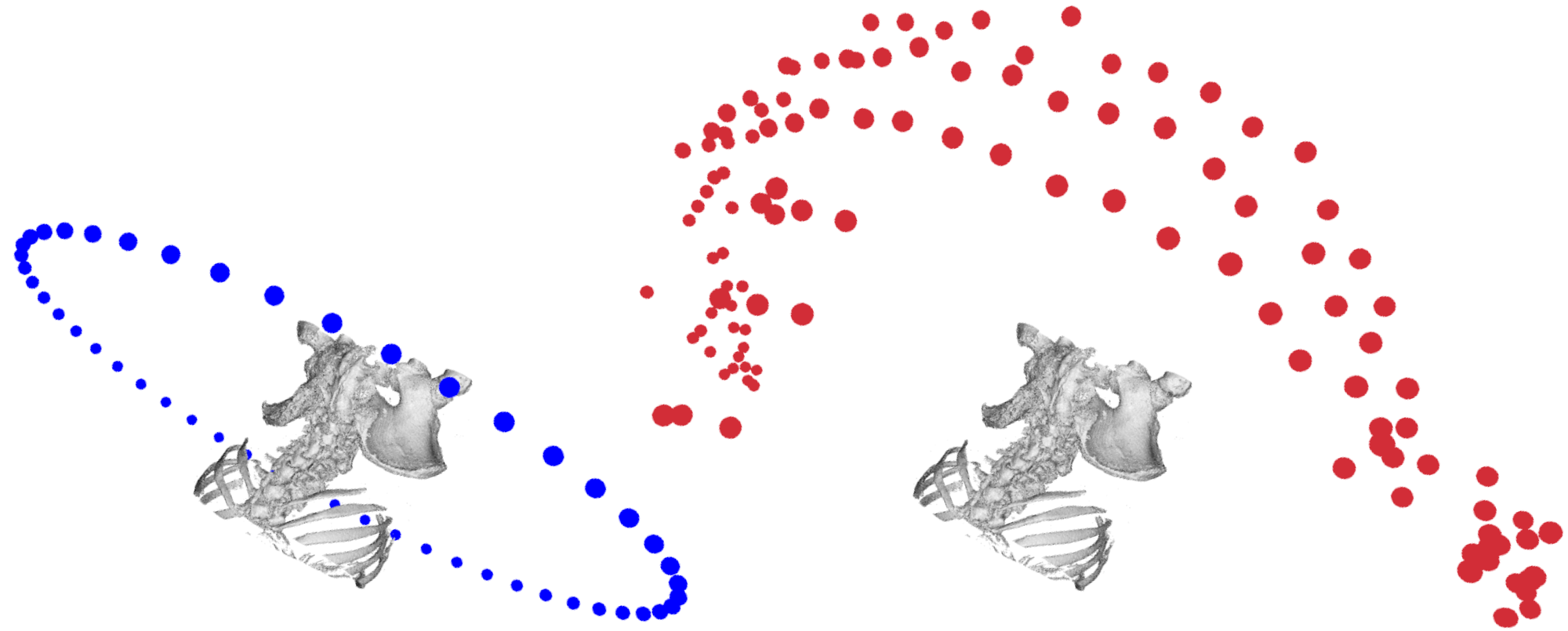}
        \caption{Comparison between conventional circular acquisition paths, typical for CBCT/CT imaging or idealized synthetic DRR generation (left), and the irregular, arbitrary acquisition poses representative of real intraoperative settings (right).}
        \label{fig:gaussian:camera_poses}
    \end{figure}

    Although these representation techniques show promise in medical imaging, such as X-rays and ultrasound, achieving accurate 3D density reconstruction with clinically acceptable quality remains challenging, especially under intraoperative conditions. For Gaussian splatting specifically, the $R^{2}$-Gaussian approach \cite{zha_r2-gaussian_2024} introduced a rectification strategy for an integration bias that previously limited volumetric accuracy. On realistic anatomy,
    the method was evaluated only through synthetic, digitally reconstructed radiographs 
    (DRRs), generated from a human
    torso CT. At the same time, validation on real X-rays was limited to simple,
    small objects such as pinecones and seashells, lacking the complexity and
    variability present in real intraoperative settings. Both synthetic and real
    data experiments relied on the simplification of circular acquisition trajectories. In surgical scenarios, however, methods must perform robustly with arbitrarily posed X-rays, as commonly encountered in intraoperative imaging (illustrated in Figure~\ref{fig:gaussian:camera_poses}). Imaging parameters such as kVp and mAs --- and therefore image contrast and brightness --- can vary noticeably from acquisition to acquisition. While $R^{2}$-Gaussian \cite{zha_r2-gaussian_2024} can generate reasonable novel views when trained on real X-rays, we observed (and literature confirms \cite{rangelov_impact_2024, ye_gaussian_2024}) that the quality of the reconstructed 3D volume is noticeably degraded under such real-world conditions.
    
    In this work, we extend $R^{2}$-Gaussian splatting \cite{zha_r2-gaussian_2024} to handle arbitrary poses and investigate its feasibility in generating 3D visualizations for surgical navigation from a sparse set of real, intraoperative 2D X-ray images. Real imaging conditions can vary greatly, and the images are often degraded by noise and other artifacts. To address this challenge, we introduce an anatomy-guided radiographic standardization step as a crucial preprocessing component. This step, implemented via a style transfer network \cite{isola_image--image_2018} trained on our paired \textit{ex-vivo} dataset \cite{jecklin_domain_2024}, maps real X-rays into a standardized domain, enhancing consistency and emphasizing bone structures to allow the subsequent Gaussian splatting optimization to focus more effectively. Following standardization, these processed images and their corresponding poses are used to optimize the 3D Gaussian representation. Finally, the optimized splats are converted into a volumetric density map, which undergoes thresholding and cropping before visualization via direct rendering or clinical slice views.

    Building upon this approach, our work explores the following two hypotheses:
    First, can Gaussian splatting generate 3D visualizations from sparse 2D X-rays
    that are deemed sufficiently accurate by experts for surgical navigation? Second,
    what is the minimum number of input X-rays required to produce a
    visualization of sufficient accuracy? To answer these questions, we first evaluated 
    the quality of the visualization via expert ratings from an experienced surgeon, 
    assessing whether the visualizations meet the requirements for intraoperative navigation. 
    Additionally, we assessed the quality using the established 2D image 
    similarity metrics PSNR and SSIM, allowing direct comparison to existing methods 
    that rely on synthetic and circularly captured data.

    A main contribution of this work is the proposed IXGS framework. This framework
    extends the $R^{2}$-Gaussian splatting approach to process arbitrary-posed real
    X-rays, demonstrating feasibility beyond previously evaluated circular
    acquisition paths. This was significantly aided by improving 3D volume consistency
    through a novel preprocessing step that emphasizes bony features using style
    transfer. Furthermore, the study provides guidelines regarding the number of
    views required to generate clinically acceptable reconstructions using this approach.

    We will release our code for 3D reconstruction from arbitrary X-rays, with a small dataset for testing, on \url{https://github.com/MrMonk3y/IXGS}.

\section*{Related Work}
    Related work is reviewed in the following subsections, covering sparse-view 3D
    reconstruction in Section~\ref{sec:introduction:sparse}, Gaussian splatting and
    neural radiance fields, especially in medical applications in Section~\ref{sec:introduction:gaussian},
    and preprocessing methods tackling the isolation of relevant anatomical regions
    in Section~\ref{sec:introduction:preprocessing}.

    \subsection*{Sparse-View 3D Reconstruction}
    \label{sec:introduction:sparse} In contrast to CT and CBCT, which generate
    high-quality 3D reconstructions from densely sampled sweeps, sparse-view approaches
    aim to achieve similar results with fewer projections. Several prior works
    have proposed methods to reconstruct 3D volumes from sparse X-ray views, each
    differing in anatomical focus and reconstruction quality. Kasten et al. \cite{kasten_end--end_2020}
    presented a U-net architecture that uses only two orthogonal X-rays of the knee
    anatomy, achieving a $90\%$ F1 score. Their approach primarily relied on synthetic
    training data, using DRRs generated
    from CT scans. To apply the trained model to real X-rays, they proposed an
    unpaired style transfer network to adapt real intraoperative X-rays into the
    appearance domain of the synthetic training data. Shiode et al.
    \cite{shiode_2d3d_2021} proposed a similar approach for the forearm, using a
    single X-ray image. They trained a Pix2Pix network on a small paired dataset
    of DRRs created through intensity-based registration of real X-rays. Although
    effective for specific anatomies, these methods rely on preoperative, diagnostic
    X-rays with near-orthogonal views, which are difficult to obtain
    consistently in realistic intraoperative settings. Another limitation is that
    end-to-end trained methods learn a priori knowledge, struggling to
    reconstruct patient-specific variations or pathologies.

    \cite{ge_x-ctrsnet_2022} presented X-CTRSNet, which reconstructed cervical vertebrae
    from two orthogonal views, achieving an $80.4\%$ F1 score. Their method handles
    real X-rays but requires standardized, well-aligned orthogonal projections
    and is limited to a small reconstruction volume of $128^{3}$ voxels. In our
    own X23D approach \cite{jecklin_x23dintraoperative_2022,jecklin_domain_2024},
    we adapt a learnt stereo machine \cite{kar_learning_2017} to reconstruct
    the lumbar spine from four arbitrarily posed intraoperative X-rays, reaching
    $84\%$ F1 score on real data. This two-stage method combined a style transfer
    network trained to bridge the synthetic-real domain gap, followed by a reconstruction
    network extensively trained on synthetic data. X23D showed promising
    accuracy and performance \cite{luchmann_spinal_2024}.

    Overall, existing sparse-view approaches demand substantial pretraining on anatomical
    datasets, making them prone to overfitting and limited in handling unseen
    pathologies or new anatomies. They also rely on voxelized outputs, which
    impose a trade-off between reconstruction volume size and resolution. In practice,
    capturing an entire bone (e.g., a femur) requires larger volumes, which can
    compromise the resolution of fine structural details like fracture lines. Finally,
    current sparse-view reconstructions lack calibrated Hounsfield units, resulting
    in density values that may not match those seen in traditional CT.

    \subsection*{Gaussian Splatting and Neural Radiance Fields}
    \label{sec:introduction:gaussian} Recent approaches for generating 3D scenes
    from images without extensive pretraining include neural radiance fields (NeRF)
    \cite{mildenhall_nerf_2021} and Gaussian splatting \cite{kerbl_3d_2023}.
    While NeRF typically uses implicit neural networks to represent scene
    density and color as a continuous field, Gaussian splatting uses an explicit
    representation. This representation displays the scene as a collection of 3D
    Gaussian primitives defined by properties such as position, covariance,
    opacity, and appearance coefficients. This representation allows Gaussian
    splatting to offer memory and computational benefits compared to NeRF-based
    approaches. Although both approaches were originally developed for novel view
    synthesis, their volumetric representations can also be converted into
    meshes for surface reconstruction \cite{rakotosaona_nerfmeshing_2024}. Both techniques
    and synthetic X-ray-based reconstructions
    \cite{corona-figueroa_mednerf_2022,cai_radiative_2025,gao_ddgs-ct_2024,wysocki_ultra-nerf_2024,awojoyogbe_neural_2024,yang_deform3dgs_2024}.

    The authors of SAX-NeRF \cite{cai_structure-aware_2024} were among the first
    to adapt NeRF for X-ray visualization by encoding density values instead of
    color and opacity. They introduced a line segment-based transformer and a ray
    sampling strategy to extract contextual and geometric information from the 2D
    projections. Although SAX-NeRF improved density reconstruction, it requires more
    than 13 hours to train on just 25 images \cite{zha_r2-gaussian_2024}.

    $R^{2}$-Gaussian \cite{zha_r2-gaussian_2024} built on Gaussian splatting to reduce
    training time to under 3 minutes for 25 images. $R^{2}$-Gaussian also
    introduced an integration bias rectification technique that significantly
    enhanced volumetric reconstruction quality. The method was tested on 25, 50,
    and 75 synthetic torso DRRs generated from circular sweeps and on real X-rays
    of relatively simple objects (e.g., walnuts, pine cones) from the FIPS X-ray
    Tomographic dataset \cite{noauthor_x-ray_nodate}.

    When applied to real intraoperative X-rays, the quality of the reconstructed
    3D volume degraded substantially in our experiments, due to more complex
    anatomy and larger imaging variations. Literature confirmed our observations.
    The complexity of the underlying topology and variations in imaging conditions
    can significantly impact the quality of the 3D volume. Previous work \cite{rangelov_impact_2024,
    ye_gaussian_2024} has shown that factors like brightness variations, noise, and
    artifacts pose severe challenges to 3D Gaussian splatting.

    To address these challenges associated with real X-rays, we developed the
    IXGS framework. This approach extends $R^{2}$-Gaussian splatting to process arbitrarily
    posed images and introduces an anatomy-guided radiographic standardization step.

    \subsection*{Preprocessing Methods}
    \label{sec:introduction:preprocessing} Several preprocessing concepts have been
    proposed to spatially isolate relevant anatomical regions and enhance their visualization
    quality. These techniques are particularly useful when working with real
    intraoperative X-rays, where noise, artifacts, and overlying or adjacent tissues
    complicate reconstruction.

    One possible approach is segmentation, where only relevant structures such
    as bones are preserved, while soft tissues and background areas are masked.
    This simplifies the reconstruction task by guiding the network on the
    structures of interest. State-of-the-art segmentation networks tailored to bone
    and spine anatomy have shown promising results \cite{chen_vertxnet_2024,kim_automatic_2021}.
    However, training segmentation networks requires extensive manual annotation
    to create labeled datasets, which is costly and time-consuming, especially for
    intraoperative imaging. Recently, pre-trained vision transformer models such
    as DINOv2 \cite{darcet_vision_2024} and SAM \cite{kirillov_segment_2023} have
    demonstrated strong performance in object segmentation, even without
    explicit annotations. To further improve segmentation performance for medical
    data, specialized adaptations such as Medical SAM have been developed
    \cite{wu_medical_2023}. Despite these advances, segmentation methods still
    struggle with low contrast, overlapping structures, and anatomical
    variability, especially in intraoperative X-rays, limiting their robustness
    in real surgical environments \cite{ehab_unet_2024}.

    An alternative is localization, which identifies the approximate region of interest
    and crops the X-ray accordingly, similar to collimation in traditional
    radiography. Localization requires less precision than full segmentation and
    can be performed using lightweight neural networks. Beyond general-purpose
    localization networks \cite{gupta_vitol_2022}, approaches trained and tailored
    to medical imaging applications have been proposed \cite{yookwan_coarse_2023,ye_projective-geometry-aware_2025}.
    In our previous work, we used ground truth X-ray poses and 3D CT segmentation
    to project individual lumbar vertebrae into the X-ray images. These
    projections were then used to train a YOLOv7 network \cite{wang_yolov7_2023}
    to localize and crop the corresponding vertebrae in real X-rays
    \cite{jecklin_domain_2024}.

    In addition to spatial isolation, appearance-based preprocessing techniques such
    as style transfer can enhance the visibility of critical structures and normalize
    image appearance across different views. Style transfer has been widely used
    to change the appearance in certain aspects, particularly in sparse-view reconstruction
    pipelines with synthetic training data \cite{kasten_end--end_2020,shiode_2d3d_2021,jecklin_domain_2024}.
    Paired style transfer methods, such as Pix2Pix \cite{isola_image--image_2018},
    leverage these paired synthetic-real datasets to train a direct mapping
    between domains. Unpaired methods, such as CycleGAN \cite{zhu_unpaired_2017},
    learn this mapping without requiring exact correspondence, making them more
    flexible when paired data is unavailable.

\section*{Methods}
    \begin{figure}
        \centering
        \includegraphics[width=0.6\textwidth]{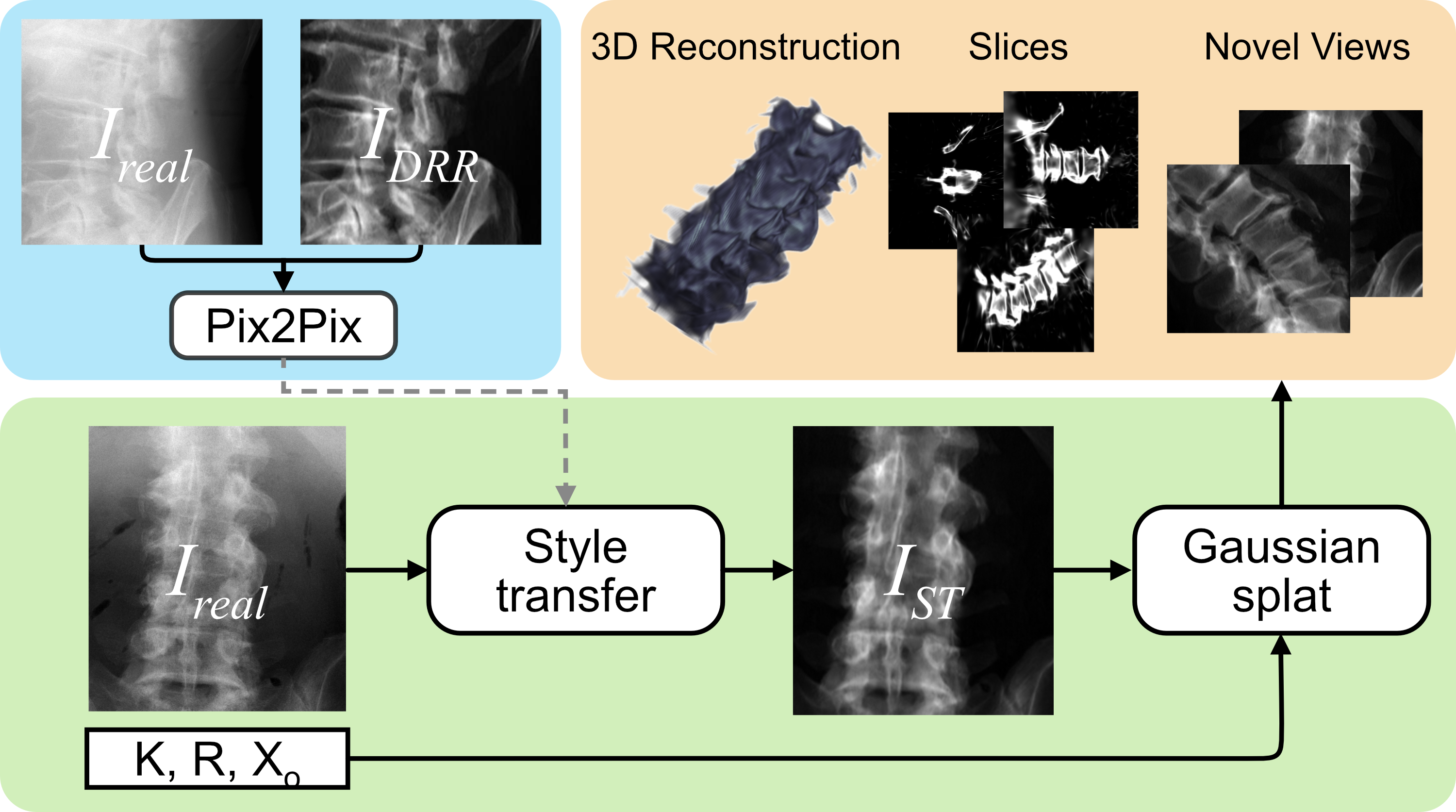}
        \caption{Pipeline Overview. The training of Pix2Pix (blue) uses paired real
        X-rays ($\mathbf{I}_{\text{real}}$) and synthetic DRRs ($\mathbf{I}_{\text{DRR}}$)
        to learn style transfer. During inference (green), real X-rays with calibration
        information are converted to style-transferred images ($\mathbf{I}_{\text{ST}}$).
        These images, along with their poses, are passed to the Gaussian
        splatting network, which outputs 3D reconstructions. The resulting
        volume can be visualized as slices or rendered from arbitrary viewpoints
        (orange).}
        \label{fig:gaussian:pipeline}
    \end{figure}

    This section details our methodology for the high-quality 3D reconstruction of
    the spine anatomy from sparse and arbitrarily posed intraoperative X-rays. An overview of the pipeline is shown in Figure~\ref{fig:gaussian:pipeline}.
    The process begins with acquiring and calibrating \textit{ex-vivo} CT and X-ray
    data to generate a paired real-synthetic dataset for training and validation
    (Sections~\ref{sec:method:data_collection}). We then employ anatomy-guided radiographic
    standardization, using a Pix2Pix model \cite{isola_image--image_2018}
    trained on this paired data, to transform real X-rays into a standardized domain
    (Section~\ref{sec:anatomy-guided-radiographic-standardization}).
    Subsequently, geometric preprocessing, including image cropping and volume normalization
    is applied to ensure compatibility with the reconstruction network (Section~\ref{sec:method:cropping_normalization}).
    Following this, volumetric 3D reconstruction is performed on these prepared views
    using our IXGS framework (Section~\ref{sec:method:gaussian}).
    The performance of this pipeline is then evaluated through experiments
    detailed in Section~\ref{sec:method:performance}, which include a qualitative
    assessment of 3D clinical usability by an expert surgeon and quantitative benchmarking
    of 2D view synthesis quality using standard metrics against baseline conditions.

    \subsection*{Dataset}
    \label{sec:method:data_collection} This study utilizes a paired dataset comprising
    spatially aligned real X-ray images $\mathbf{I}_{\text{real}}$ and corresponding
    synthetic DRRs $\mathbf{I}_{\text{DRR}}$ generated from CT scans. This
    specialized dataset served two critical purposes within our pipeline: first,
    it provided the necessary aligned image pairs for training the Pix2Pix
    network used in the anatomy-guided radiographic standardization step (Section~\ref{sec:anatomy-guided-radiographic-standardization}).
    Second, it served as ground truth data and a defined target
    domain for benchmarking reconstruction performance and evaluating the
    effectiveness of the standardization process (Section~\ref{sec:method:performance}).
    The detailed construction of this dataset, including calibration and pairing,
    has been described in detail in our previous work \cite{jecklin_domain_2024}.
    The data collection process can be summarized as follows.

    We constructed a data collection pipeline to acquire such a paired \textit{ex-vivo}
    dataset consisting of real $\mathbf{I}_{\text{real}}$ and $\mathbf{I}_{\text{DRR}}$
    images captured from the same arbitrary viewpoints, along with their
    corresponding pose information. The study was conducted according to the guidelines of the Declaration of Helsinki and approved by the local ethical committee (KEK Zurich BASEC No. 2021-01083). The CT data was obtained using a SOMATOM Edge
    Plus system (Siemens Healthcare, Erlangen, Germany), providing 0.75 mm slice
    thickness and 0.5 mm $\times$ 0.5 mm in-plane resolution. Subsequently,
    segmentations of the L1 to L5 vertebrae within these CT volumes were
    generated by trained specialists employing Mimics software (Version 25.0,
    Materialise NV, Leuven, Belgium). Although the reconstructed 3D volumes in
    this work do not exhibit clear surface boundaries suitable for direct mesh
    extraction, these segmentations served as ground truth references during qualitative
    evaluation.

    X-rays were acquired from anterior-posterior (AP), lateral, and oblique views
    using a clinical-grade mobile C-arm (Cios Spin®, Siemens Healthineers,
    Erlangen, Germany). The C-arm was rotated in 3° increments over a transverse
    orbit of ±102° and tilted in 3° increments over a sagittal range of ±25°.
    Additional lateral views were captured with 5° sagittal tilts over a range
    of ±15°. The X-ray beam was centered on the L3 vertebra to ensure full coverage
    of the lumbar spine.

    To achieve accurate pose information, 14 stainless steel spherical fiducials
    were inserted into the soft tissue surrounding the spinal column from L1 to
    L5. A post-fiducial placement CT scan was acquired to determine the 3D positions
    of the fiducials, which were identified through thresholding and 3D
    connected component analysis.

    The calibration process involved detecting fiducial projections in the X-rays
    using a circular Hough transform \cite{duda_use_1972}, followed by user verification
    if necessary. Initial pose estimates were obtained using the direct linear
    transform (DLT) algorithm \cite{abdel-aziz_direct_2015}, refined with a RANSAC-based
    process to minimize reprojection error. Once optimal correspondences were
    established, the final camera matrix $\mathsf{P}$ was estimated using all fiducials.
    This matrix was then decomposed into intrinsic parameters $\mathsf{K}$, the rotation
    matrix $\mathsf{R}$, and the camera origin $\mathbf{X}_{o}$. To preserve realism
    in the dataset, fiducial projections were subsequently inpainted from the X-rays.

    Following this, for each real X-ray $\mathbf{I}_{\text{real}}$, a synthetic
    $\mathbf{I}_{\text{DRR}}$ was generated from the corresponding CT scan and
    calibration information. The transformation from world space to C-arm space
    is defined by the matrix
    \begin{equation}
        \mathsf{T}=
        \begin{bmatrix}
            \mathsf{R}        & \mathbf{X}_{o} \\
            \mathbf{0}^{\top} & 1
        \end{bmatrix}, \label{eq:T}
    \end{equation}

    which incorporates the extrinsic camera parameters. The DRR at pixel $p$ is computed
    by:

    \begin{equation}
        \mathbf{I}_{\text{DRR}}(p) = \int A\left(\mathsf{T}^{-1}L(p, s)\right) \,
        \mathbf{1}\left ( A\left( \mathsf{T}^{-1}L(p, s) \right) > t \right) \, d
        s
    \end{equation}

    where $A$ denotes the CT attenuation values, $L(p, s)$ is the ray passing through
    pixel $p$ parameterized by depth $s$, and $\mathbf{1}(\cdot)$ is an
    indicator function returning $1$ if the attenuation exceeds a threshold $t$. For
    this study, we set $t = 0$ Hounsfield units to isolate bone structures.

    This study utilized six \textit{ex-vivo} human specimens selected from the original
    dataset created in \cite{jecklin_domain_2024}. The selection ensured that
    each specimen had at least 50 X-rays available for training, along with a sufficient
    number of images for testing. For each \textit{ex-vivo} specimen, 50 X-rays
    were randomly sampled to form the training set, encompassing a diverse range
    of views including anterior-posterior, lateral, oblique, and off-axis shots.
    This resulted in a total of 300 training images. The remaining X-rays were
    reserved as dedicated test sets for each specimen, resulting in a combined test
    set of 233 images. Notably, the same test set was used consistently across all
    experiments, regardless of the number of training views.

    \subsection*{Anatomy-guided Radiographic Standardization}
    \label{sec:anatomy-guided-radiographic-standardization}

    To ensure consistent input quality and emphasize key anatomical structures,
    we applied an anatomy-guided radiographic standardization step before
    feeding the images to the Gaussian splatting network. This preprocessing step
    aims to isolate the region of interest, enhance bone structures, and reduce variability
    in appearance caused by, e.g., noise and artifacts present in real X-rays. By
    mapping real X-ray images to a standardized synthetic domain, we improve the
    robustness of the subsequent volumetric reconstruction process.

    The paired real X-ray images $\mathbf{I}_{\text{real}}$ and synthetic DRRs
    $\mathbf{I}_{\text{DRR}}$ were used to train a Pix2Pix model
    \cite{isola_image--image_2018}. The objective of the generative adversarial network
    (GAN) is to map $\mathbf{I}_{\text{real}}$ to the domain of $\mathbf{I}_{\text{DRR}}$,
    emphasizing bone structures and ensuring a consistent visual appearance
    across all views. This standardization process is critical for enabling the Gaussian
    splatting network to effectively focus on relevant anatomical features necessary
    for accurate 3D reconstruction.

    The GAN consists of a generator $G$ and a discriminator $D$. These are
    optimized using a combination of adversarial loss
    $\mathcal{L}_{\text{cGAN}}$ and L1 loss $\mathcal{L}_{L1}$ to improve image
    quality and reduce blurring, defined as follows:

    \begin{align}
        \mathcal{L}_{\text{cGAN}}(G,D) & = \mathbb{E}_{\mathbf{I}_{\text{real}},\mathbf{I}_{\text{DRR}}}[\log D(\mathbf{I}_{\text{real}},\mathbf{I}_{\text{DRR}})] \nonumber                           \\
                                       & \quad + \mathbb{E}_{\mathbf{I}_{\text{real}},z}[\log(1-D(\mathbf{I}_{\text{real}},G(\mathbf{I}_{\text{real}},z)))] \label{eq:pix2pix_gan_loss}                \\
        \mathcal{L}_{L1}(G)            & = \mathbb{E}_{\mathbf{I}_{\text{real}},\mathbf{I}_{\text{DRR}},z}[\|\mathbf{I}_{\text{DRR}}- G(\mathbf{I}_{\text{real}}, z)\|_{1}] \label{eq:pix2pix_l1_loss} \\
        G^{*}                          & = \arg\min_{G}\max_{D}\mathcal{L}_{\text{cGAN}}(G,D) + \lambda\mathcal{L}_{L1}(G) \label{eq:pix2pix_objective}
    \end{align}

    Here, the variable $z$ denotes random noise input to the generator $G$,
    typically introduced via dropout, allowing for stochasticity in
    the generated output $G(\mathbf{I}_{\text{real}}, z)$.

    To prevent information leakage, a separate Pix2Pix model was trained for
    each \textit{ex-vivo} specimen, ensuring that each test specimen remained
    completely excluded from its corresponding training set. After training, the
    generator $G$ was used to style-transfer $\mathbf{I}_{\text{real}}$ to the
    synthetic domain, producing $\mathbf{I}_{\text{ST}}= G(\mathbf{I}_{\text{real}}
    )$.

    \subsection*{Preprocessing: Cropping and Volume Normalization}
    \label{sec:method:cropping_normalization} After anatomy-guided radiographic standardization,
    additional preprocessing steps were applied to all synthetic and real images
    to ensure compatibility with the Gaussian splatting network.

    Due to slight mechanical deflection of the C-arm under gravitational load, the
    principal point $(c_{x}, c_{y})$ derived from the intrinsic matrix $\mathsf{K}$
    during calibration is often not centered in the image. Since Gaussian
    splatting assumes a centered principal point; we cropped the images around $(
    c_{x}, c_{y})$ to ensure alignment within a fixed resolution of $512 \times 5
    12$ pixels.

    Furthermore, the 3D volume to be reconstructed via Gaussian splatting was
    normalized to fit within the spatial range $[-1, 1]^{3}$. This normalization
    is required by Gaussian splatting frameworks to ensure consistent handling of
    spatial coordinates during training and inference.
    
    \subsection*{3D Reconstruction via Gaussian Splatting}
    \label{sec:method:gaussian} 
    Our IXGS approach builds upon the $R^{2}$-Gaussian splatting framework introduced by Zha et al.~\cite{zha_r2-gaussian_2024}. The method represents the target anatomical structure by a collection of learnable Gaussian kernels $\mathbb{G}^{3}= \{G^{3}_{j}\}_{j=1,\ldots,M}$. Each Gaussian kernel $G^{3}_{j}$ describes a local density distribution in 3D space:
    \begin{equation}
        G^{3}_{j}(\mathbf{x}) = \rho_{j}\exp\left( -\frac{1}{2}(\mathbf{x}- \mathbf{p}_{j})^{\top}\mathbf{\Sigma}_{j}^{-1}(\mathbf{x}- \mathbf{p}_{j}) \right),
        \label{eq:gaussian_kernel} 
    \end{equation}
    where $\mathbf{x}\in \mathbb{R}^{3}$ is a point in world coordinates, $\rho_{j}\in \mathbb{R}$ represents the central density, $\mathbf{p}_{j}\in \mathbb{R}^{3}$ the position of the kernel, and $\mathbf{\Sigma}_{j}\in \mathbb{R}^{3 \times 3}$ its covariance matrix. Unlike standard 3D Gaussian splatting methods, the \cite{zha_r2-gaussian_2024} approach omits view-dependent color information since X-ray imaging is exclusively based on isotropic density.
    
    \paragraph{Initialization}
    In contrast to $R^{2}$-Gaussian splatting \cite{zha_r2-gaussian_2024}, which
    initializes kernels with the Feldkamp-Davis-Kress (FDK) algorithm
    \cite{feldkamp_practical_1984}, our IXGS approach requires random initialization
    due to the non-circular and arbitrary acquisition poses used in our pipeline,
    where FDK is not applicable. For the evaluations presented in this work,
    where ground truth CT data was available, the center and dimensions of the reconstruction
    volume were defined based on the CT. For future clinical applications, the volume
    origin/dimensions would need to be estimated directly from the calibrated input
    views, for instance, using techniques like the linear triangulation of principal
    image axes demonstrated in our prior work \cite{jecklin_domain_2024}.
    Initial kernel positions $\mathbf{p}_{i}$ were then randomly sampled within
    this volume. Initial densities $\rho_{i}$ are uniformly sampled, and the
    covariances $\mathbf{\Sigma}_{i}$ are initialized as isotropic matrices with
    predefined scale values.

    \paragraph{Projection and Rendering}
    We refine the 3D Gaussian kernels $\mathbb{G}^{3}$ by matching their rendered
    projections against a set of $N$ input 2D images $\{\mathbf{I}^{i}\}_{i=1..N}$
    (representing $\mathbf{I}_{\text{ST}}$, $\mathbf{I}_{\text{real}}$, or
    $\mathbf{I}_{\text{DRR}}$ depending on the experiment) acquired from known camera
    poses $\{\mathsf{P}^{i}\}_{i=1..N}$. For each input view $i$, a corresponding
    projected image $I_{\text{proj}}^{i}$ is generated by rendering the current 3D
    Gaussians $\mathbb{G}^{3}$ using pose $\mathsf{P}^{i}$.

    Conceptually, the value of a pixel in $I_{\text{proj}}^{i}$ corresponding to
    a camera ray $\mathbf{r}(t)$ is computed by integrating the densities of all
    3D Gaussian kernels $G^{3}_{j}$ along that ray:
    \begin{equation}
        I_{\text{proj}}^{i}(\mathbf{r}) = \sum_{j=1}^{M}\int G^{3}_{j}(\mathbf{r}
        (t)) \, dt. \label{eq:projection_integral}
    \end{equation}
    Here, $G^{3}_{j}(\mathbf{r}(t))$ (defined by $\mathbf{p}_{j}, \mathbf{\Sigma}
    _{j}, \rho_{j}$) refers to the density function evaluated along the ray r(t),
    which originates from the X-ray source and intersects the image plane at
    specific pixel coordinates.

    To project the Gaussian kernels for X-ray rendering, we first map the 3D Gaussians
    ($G^{3}_{j}$) from world space to the C-arm space specific to view $i$ using
    the transformation $\mathsf{T}^{i}$ (Equation~\ref{eq:T}). In the C-arm space,
    the origin is at the X-ray source, and the z-axis points toward the projection
    center.

    Since a C-arm with its cone-beam X-ray scanner can be modeled similarly to a pinhole camera,
    we follow \cite{zwicker_ewa_2002,zha_r2-gaussian_2024} and further transfer
    the Gaussians from C-arm space to a ray space. In ray space, the viewing
    rays are aligned parallel to the third coordinate axis, simplifying the geometry and facilitating analytical integration along the ray's path. Due to
    the non-Cartesian nature of ray space, a local affine transformation is applied
    \cite{zwicker_ewa_2002,zha_r2-gaussian_2024}. In this step, the Jacobian of the
    affine approximation is computed from the full projection --- which, when linearized
    at the kernel centers, implicitly incorporates the shot-specific intrinsic
    parameters --- and is used to scale the original Gaussian parameters. This
    yields view-dependent 3D parameters in ray space: position $\tilde{\mathbf{p}}
    _{j}^{i}$ and covariance $\tilde{\mathbf{\Sigma}}_{j}^{i}$ for each 3D Gaussian
    $j$ as observed from view $i$.

    Following the derivation in \cite{zha_r2-gaussian_2024}, the integration
    along the ray-space viewing axis ($x_{2}$) results in a sum of 2D Gaussians
    projected onto the image plane. Let $\hat{\mathbf{x}}\in \mathbb{R}^{2}$ be the
    2D coordinates on the image plane corresponding to ray $\mathbf{r}$. Let
    $\hat{\mathbf{p}}_{j}^{i} \in \mathbb{R}^{2}$ and
    $\hat{\mathbf{\Sigma}}_{j}^{i} \in \mathbb{R}^{2\times2}$ be the 2D position
    and the covariance obtained by dropping the third row and column from
    $\tilde{\mathbf{p}}_{j}^{i}$ and $\tilde{\mathbf{\Sigma}}_{j}^{i}$, respectively.
    To ensure accurate density reconstruction across views, the original 3D density
    $\rho_{j}$ is scaled by a view-dependent, covariance-related factor
    $\mu_{j}^{i} = \sqrt{2\pi|\tilde{\mathbf{\Sigma}}_{j}^{i}| / |\hat{\mathbf{\Sigma}}_{j}^{i}|}$,
    yielding the effective 2D density $\hat{\rho}_{j}^{i} = \mu_{j}^{i} \rho_{j}$.
    This scaling corrects the integration bias inherent in standard Gaussian
    splatting projections, as detailed in \cite{zha_r2-gaussian_2024}. The final
    rendered intensity at pixel $\hat{\mathbf{x}}$ is then computed by summing
    these projected 2D Gaussian contributions:
    \begin{equation}
        I_{\text{proj}}^{i}(\hat{\mathbf{x}}) \approx \sum_{j=1}^{M}G^{2}_{j}(\hat
        {\mathbf{x}}|\hat{\rho}_{j}^{i}, \hat{\mathbf{p}}_{j}^{i}, \hat{\mathbf{\Sigma}}
        _{j}^{i}), \label{eq:projection_final_sum}
    \end{equation}
    where $G^{2}_{j}$ represents the contribution of the $j$-th Gaussian projected
    onto the 2D image plane, parameterized by its effective 2D density
    $\hat{\rho}_{j}^{i}$, position $\hat{\mathbf{p}}_{j}^{i}$, and covariance
    $\hat{\mathbf{\Sigma}}_{j}^{i}$. Unlike the original $R^{2}$-Gaussian work,
    which focused on circular paths, our IXGS approach implements this projection
    for arbitrary poses $\{\mathsf{P}^{i}\}_{i=1..N}$. The resulting rendered
    projections $\{I_{\text{proj}}^{i}\}_{i=1..N}$ provide the basis for
    optimizing the 3D scene representation.

    \paragraph{Optimization}
    The optimize the 3D Gaussian kernels ($\mathbb{G}^{3}$) we minimize a loss function
    calculated by comparing each rendered projection $I_{\text{proj}}^{i}$ to its
    corresponding input image $\mathbf{I}^{i}$ and summing the loss across all $N$
    views. The loss combines an L1 difference, the structural similarity index measure
    (SSIM) \cite{wang_image_2004}, and total variation
    \cite{rudin_nonlinear_1992}. The derived gradients of this total loss are then
    backpropagated to adjust the Gaussian parameters ($\mathbf{p}_{j}$, $\mathbf{\Sigma}
    _{j}$, $\rho_{j}$). This optimization, in combination with adaptive density control
    mechanisms adopted from \cite{kerbl_3d_2023, zha_r2-gaussian_2024},
    iteratively refines the 3D Gaussians ($\mathbb{G}^{3}$) until they represent
    an accurate 3D scene consistent with the input sparse views.

    \paragraph{Output}
    For the final 3D reconstruction, the optimized Gaussian kernels
    $\mathbb{G}^{3}$ are converted into a volumetric density map $\mathbf{V}$
    using the differentiable voxelizer operator $\mathcal{V}$ from
    \cite{zha_r2-gaussian_2024}:
    \begin{equation}
        \mathbf{V}= \mathcal{V}(\mathbb{G}^{3}) \label{eq:V}
    \end{equation}
    This operator partitions the target space into discrete 3D tiles and aggregates
    kernel contributions within each tile.

    Using Equation~\ref{eq:V}, we generate volumes (denoted $\mathbf{V}_{ST}$ or
    $\mathbf{V}_{real}$ depending on the input image type) which predominantly
    capture bony structures, alongside some soft tissue and occasional artifacts
    inherent to the splatting process. To suppress artifacts and isolate the
    region of interest (e.g., the lumbar spine), postprocessing involving a thresholding
    step (empirically set to the 80th percentile) and cropping is applied.

    These processed volumes ($\mathbf{V}$) are the primary 3D output used for qualitative
    evaluation. They can be visualized directly using 3D rendering techniques. To
    further enhance inspection of the 3D volume, interactive clipping planes, oriented
    parallel to the volume faces, were implemented. These movable planes allow users
    to cut through the volume and reveal internal structures. Alternatively, the
    volumes can be inspected via standard clinical multiplanar slices (axial, coronal,
    and sagittal).

    Additionally, for quantitative benchmarking purposes, 2D projections $\mathbf{I}
    _{\text{proj}}$ were generated by rendering the optimized 3D Gaussians ($\mathbb{G}
    ^{3}$) from the poses corresponding to the held-out test set images (using
    Equation~\ref{eq:projection_final_sum}).

    \subsection*{Experiments and Performance Evaluation}
    \label{sec:method:performance} The performance evaluation assessed the reconstructed
    3D volumes and the quality of novel synthesized 2D views. The 3D evaluation
    focused on a qualitative assessment of clinical usability performed by an
    expert surgeon. The 2D evaluation used standard quantitative metrics (peak signal-to-noise ratio (PSNR), SSIM)
    calculated by comparing synthesized views against held-out test images, for benchmarking
    against prior work.

    \paragraph{3D Evaluation}
    \begin{figure}
        \centering
        \includegraphics[width=1.0\textwidth]{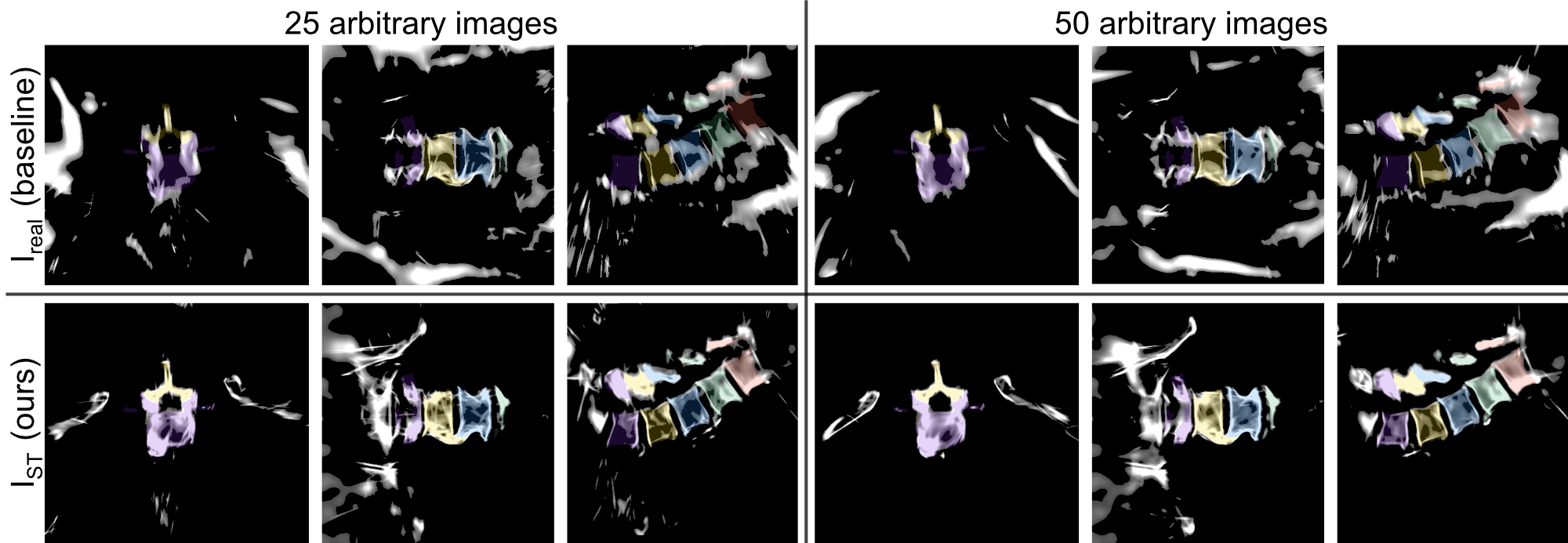}
        \caption{Comparison of 3D reconstructions of the lumbar spine (axial,
        coronal, and sagittal slices). Each block shows axial, coronal, and sagittal
        slices of the reconstructed volume, overlaid with alpha-blended masks of
        the segmented lumbar spine from the ground truth CT for better accuracy
        assessment. The columns compare reconstructions using 25 views (left) and
        50 views (right). The top row shows reconstructions from
        $\mathbf{I}_{\text{real}}$ baseline, while the bottom row displays reconstructions
        from $\mathbf{I}_{\text{ST}}$ (our approach).}
        \label{fig:gaussian:slices}
    \end{figure}
    An experienced orthopedic surgeon reviewed the 3D reconstructions to rate
    their anatomical accuracy and usability for intraoperative navigation tasks,
    specifically focusing on pedicle screw placement. The evaluation centered on
    clearly depicting key lumbar spine landmarks, including the pedicles,
    spinous processes, and vertebral endplates. Ratings were provided using the
    following 4-point Likert scale:
    \begin{itemize}
        \item[4:] \textbf{Very good} --- Clear anatomy and reliable orientation;
            suitable for navigation.

        \item[3:] \textbf{Acceptable} --- All major landmarks discernible; acceptable
            for navigation.

        \item[2:] \textbf{Poor} --- Only some major landmarks discernible, potentially
            limiting utility for navigation.

        \item[1:] \textbf{Unusable} --- Anatomical structures indistinguishable;
            ineffective for navigation.
    \end{itemize}

    For each reconstruction, generated using varying numbers of arbitrary $\mathbf{I}
    _{\text{ST}}$ images (ranging from 5 to 50 views, in increments of 5), the surgeon
    evaluated by examining both the direct 3D visualization and the
    slices (sagittal, coronal, and axial) in two independent sessions. To aid
    the evaluation, the surgeon utilized an overlay feature displaying the ground
    truth segmentations for vertebrae L1 to L5 (see Figure~\ref{fig:gaussian:slices}).
    This allowed for direct comparison between the reconstructed anatomy slices
    and the known ground truth, enabling the surgeon to toggle the overlay freely
    to assess spatial alignment and anatomical completeness.

    \paragraph{2D Quantitative Evaluation and Projection Assessment}
    We assessed reconstruction quality quantitatively using the established 2D
    metrics PSNR and SSIM \cite{wang_image_2004}, comparing
    rendered views against held-out test images. We evaluated reconstruction
    based on $\mathbf{I}_{\text{ST}}$ against reconstructions from
    $\mathbf{I}_{\text{DRR}}$ and $\mathbf{I}_{\text{real}}$. Each of these configurations
    was tested using both 50 and 25 input images per specimen. The
    $\mathbf{I}_{\text{DRR}}$ images --- generated directly from the ground truth CT
    scans using the exact same arbitrary poses as the corresponding $\mathbf{I}_{\text{real}}$
    images --- serve as a synthetic benchmark and represent the target appearance
    domain for our proposed $\mathbf{I}_{\text{ST}}$ radiographic
    standardization step. Second, as a baseline representing idealized
    conditions, we applied the original $R^{2}$-Gaussian framework \cite{zha_r2-gaussian_2024}
    to synthetic \textit{ex-vivo} human images generated with circular
    acquisition paths referred to as ($\mathbf{I}_{\text{circ}}$), also using 50
    and 25 views per specimen.

    To investigate the impact of view count on reconstruction quality, we
    measured PSNR and SSIM while varying the number of input images for one
    representative \textit{ex-vivo} specimen. Input views were incrementally
    increased from 5 to 50 (in steps of 5), and the resulting reconstruction quality
    was evaluated against the corresponding test set.

    We also compared our results against the benchmark performance of the original
    $R^{2}$-Gaussian framework \cite{zha_r2-gaussian_2024} on real X-rays of the
    simpler FIPS dataset objects (walnuts, pinecones, seashells), which also
    utilized circular acquisition paths (using 50 and 25 views).

    These quantitative comparative experiments allow us to isolate and assess
    the effects of different acquisition paths (circular vs. arbitrary) and image
    types (synthetic, real, style-transferred), thereby evaluating the efficacy
    of our framework adaptations and the proposed radiographic standardization
    step ($\mathbf{I}_{\text{ST}}$).

    \subsection*{Implementation Details}
    \label{sec:method:implementation} For the implementation of our proposed
    IXGS framework, we adapted the official implementation of $R^{2}$-Gaussian ({\url{https://github.com/Ruyi-Zha/r2_gaussian}).
    Our adapted code will be released at \url{https://github.com/MrMonk3y/IXGS}.

    For the anatomy-guided standardization, we utilized the official PyTorch implementation
    of Pix2Pix (\url{https://github.com/junyanz/pytorch-CycleGAN-and-pix2pix})
    with default hyperparameters. Due to memory constraints, images were resized
    to $512\times512$ pixels before training. The network was trained for 200 epochs,
    with a batch size of 1. The learning rate was set to $0.0002$, and the Adam optimizer
    was used with $\beta_{1}=0.5$ and $\beta_{2}=0.999$.

    The training of the Gaussian splats was conducted on an NVIDIA A100 GPU with
    40GB of memory. We followed the default hyperparameters provided by the
    authors of $R^{2}$ Gaussian \cite{zha_r2-gaussian_2024}, which produced slightly
    better results for our target volumes based on empirical evaluation. The
    network was trained for 30k iterations. 

\section*{Results}
    \label{sec:results}
    \begin{figure}
        \centering
        \includegraphics[width=0.7\textwidth]{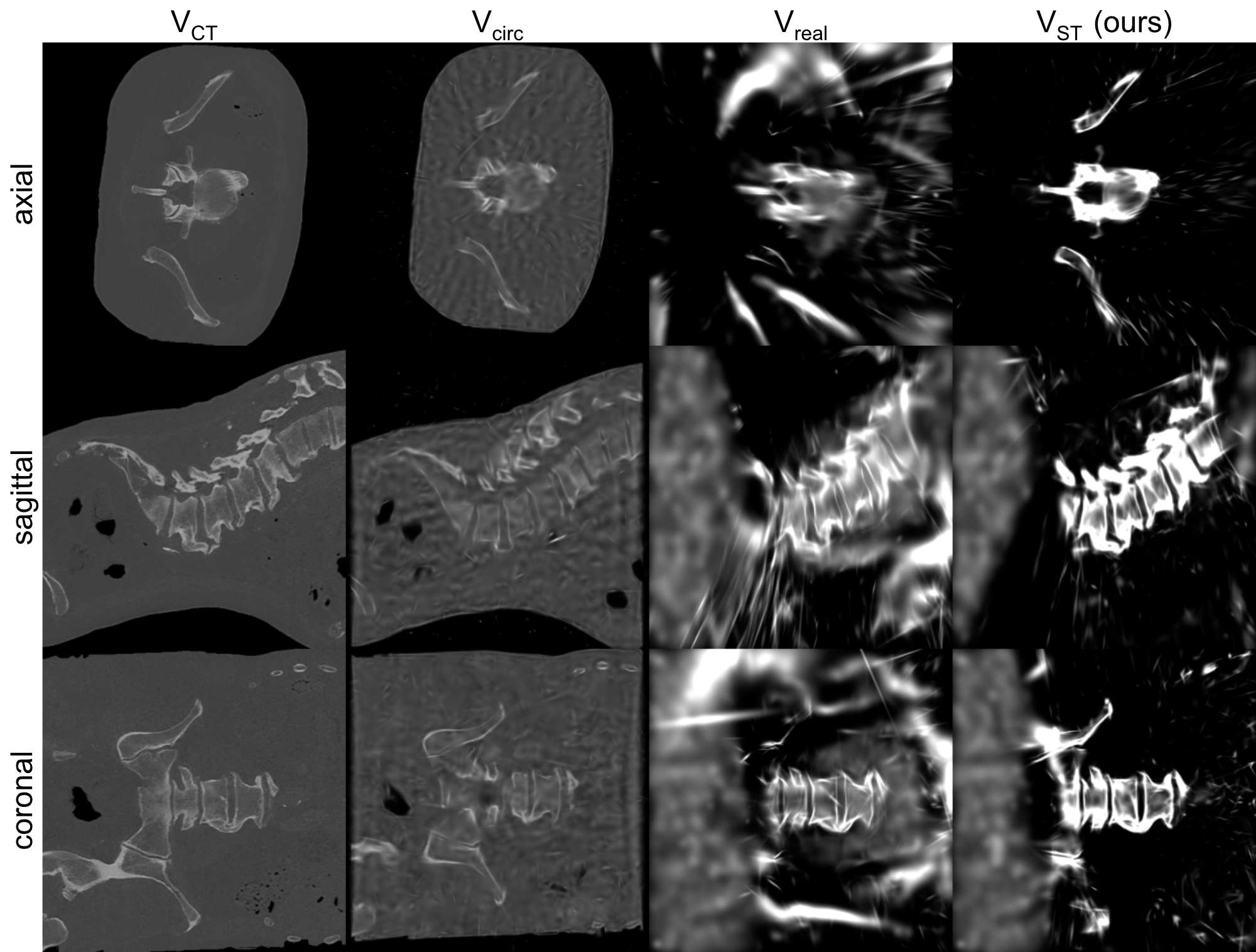}
        \caption{Comparison of slices from reconstructed volumes using different
        inputs and methods. From left to right: $\mathbf{V}_{\text{CT}}$: Ground
        Truth CT volume, $\mathbf{V}_{\text{circ}}$: Reconstruction from 50 synthetic
        DRRs generated from circular acquisition, $\mathbf{V}_{\text{real}}$: Reconstruction
        from 50 X-rays generated from arbitrary poses, $\mathbf{V}_{\text{ST}}$
        (our approach): Reconstruction from 50 style-transferred X-rays.}
        \label{fig:gaussian:comparison}
    \end{figure}
    This section details the performance evaluations conducted in this study. We
    begin with the qualitative 3D assessment, where an experienced surgeon rated
    the clinical usability of the reconstructed volumes (Section~\ref{sec:results:qualitative_3d}).
    Following this, we report the quantitative metrics assessing the 2D reconstruction
    quality across different input types, acquisition geometries, and view counts,
    including comparisons to baseline methods (Section~\ref{sec:results:quantitative}).

    Figure~\ref{fig:gaussian:comparison} provides an initial comparison of
    representative slices, illustrating the core challenges addressed in this work.
    The reconstruction from a circular acquisition path $\mathbf{V}_{\text{circ}}$
    closely resembles the ground truth CT slices $\mathbf{V}_{\text{CT}}$, demonstrating
    the potential reconstruction quality under idealized conditions. However, the
    reconstruction using $\mathbf{I}_{\text{real}}$ shows a notable decrease in quality.
    Anatomical clarity is reduced, particularly evident in the axial slice, where
    bone structures are indistinct. In contrast, the reconstruction using $\mathbf{I}
    _{\text{ST}}$ shows clear bone structures. A general characteristic visible
    in all presented Gaussian splat reconstructions is the presence of residual
    cloudy regions, which correspond to areas not covered by the input views,
    originating from the random initialization.

    \subsection*{Assessment of the 3D Volume Reconstruction Quality}
    \label{sec:results:qualitative_3d}
    \begin{figure}
        \centering
        \includegraphics[width=1.0\textwidth]{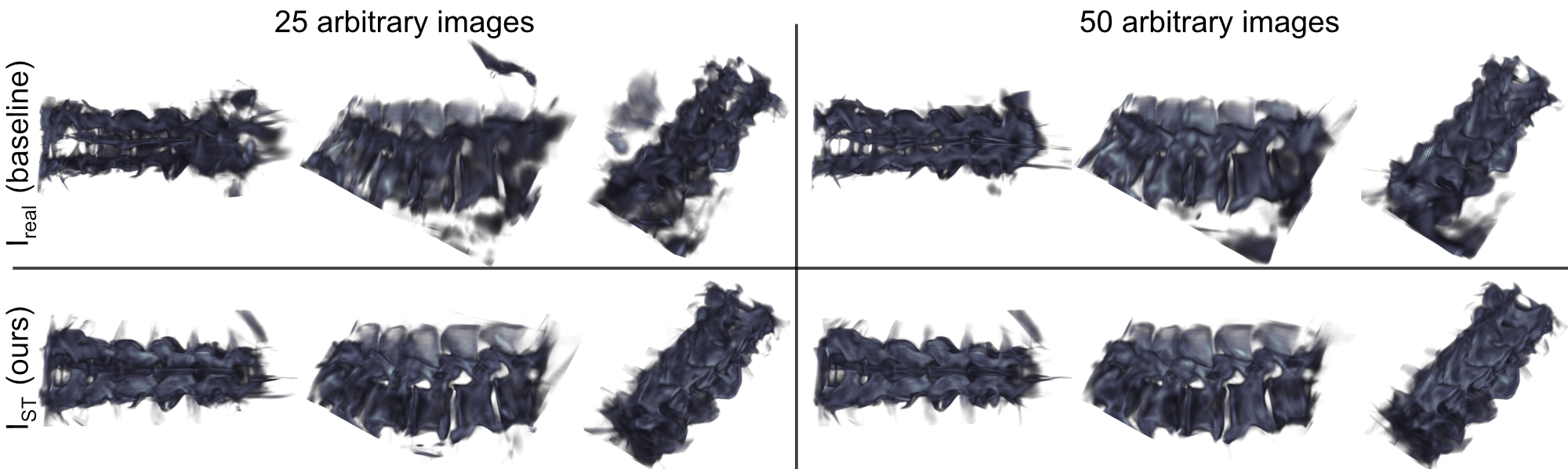}
        \caption{Comparison of 3D reconstructions of the lumbar spine. Each block
        shows AP, lateral, and isometric views of the reconstructed volume. The
        columns compare reconstructions using 25 views (left) and 50 views (right).
        The top row displays reconstructions from $\mathbf{I}_{\text{real}}$ baseline,
        while the bottom row shows reconstructions from $\mathbf{I}_{\text{ST}}$
        (our approach).}
        \label{fig:gaussian:3D}
    \end{figure}
    Figures~\ref{fig:gaussian:3D} and \ref{fig:gaussian:slices} show
    representative examples of the volumetric renderings and corresponding slice
    views, respectively, that were presented independently to the surgeon for
    evaluation. The surgeon's usability ratings, aggregated over all six \textit{ex
    vivo} specimens using 5 to 50 input $\mathbf{I}_{\text{ST}}$ images (in
    increments of 5), are summarized in Figure~\ref{fig:gaussian:ratings}a) for
    volume renderings and Figure~\ref{fig:gaussian:ratings}b) for slice views.

    For the direct volume rendering assessment (Figure~\ref{fig:gaussian:ratings}a),
    ratings were generally 'Poor' or 'Unusable' when using less than 15 input
    images. From 20 input images onwards, some specimens achieved 'Acceptable'
    ratings, with 'Very Good' ratings appearing at 30 images. Notably, one specimen
    never surpassed 'Poor' ratings regardless of the number of input images used
    for reconstruction.

    A similar trend was observed for the slice view assessment (Figure~\ref{fig:gaussian:ratings}b),
    although the quality thresholds shifted slightly. The first 'Acceptable'
    ratings for slice visualizations were achieved with 15 input images. However,
    ratings generally remained lower than those for the corresponding volume
    renderings up to 45 input images. For one specimen, slice ratings remained 'Poor'
    until 50 input images were used, at which point they reached an 'Acceptable'
    level. At the maximum of 50 input images, the majority of ratings for slice views
    indicated 'Very Good' quality.

    \begin{figure}
        \centering
        \includegraphics[width=0.7\linewidth]{
            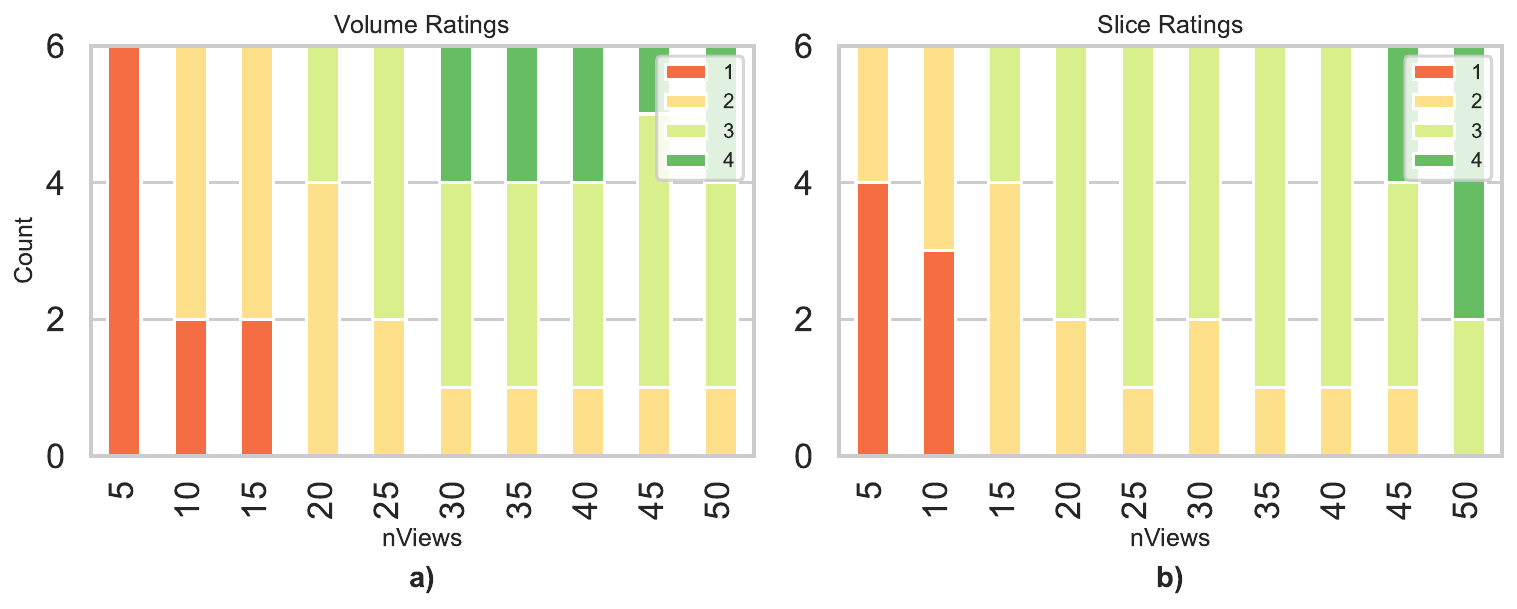
        }
        \caption{Evaluation of reconstruction quality using Likert ratings from
        expert surgeons. a) and b) show ratings over 5 to 50 views, with a) assessing
        3D volumes and b) assessing slice representations. See Section~\ref{sec:method:performance}
        for details on the different Likert scales used.}
        \label{fig:gaussian:ratings}
    \end{figure}

    \subsection*{Quantitative 2D Reconstruction Quality}
    \label{sec:results:quantitative}
    \begin{table}
        \centering
        \renewcommand{\arraystretch}{1.2}
        \setlength{\tabcolsep}{8pt}
        \begin{tabular}{c|cc|cc}
            \hline
            \multirow{2}{*}{\textbf{Method}} & \multicolumn{2}{c|}{\textbf{50 Views}} & \multicolumn{2}{c}{\textbf{25 Views}} \\
                                             & PSNR $\uparrow$                        & SSIM $\uparrow$                      & PSNR $\uparrow$ & SSIM $\uparrow$ \\
            \hline
            $I_{\mathrm{circ}}$              & 39.19                                  & 0.970                                & 33.51           & 0.940           \\
            $I_{\mathrm{DRR}}$               & 29.45                                  & 0.890                                & 26.62           & 0.830           \\
            $I_{\mathrm{real}}$              & 23.22                                  & 0.760                                & 20.52           & 0.720           \\
            $I_{\mathrm{ST}}$ (ours)         & 25.73                                  & 0.790                                & 24.17           & 0.750           \\
            \hline
        \end{tabular}
        \caption{Comparison of PSNR and SSIM for different input types. The input
        types include $\mathbf{I}_{\text{circ}}$ (a synthetic and circular
        baseline), $\mathbf{I}_{\text{DRR}}$ (synthetic DRRs), $\mathbf{I}_{\text{real}}$
        (real X-rays), and $\mathbf{I}_{\text{ST}}$ (style-transferred, ours).
        Performance is measured across two view counts: 25 and 50.}
        \label{tab:results}
    \end{table}

    Table~\ref{tab:results} compares the average PSNR and SSIM across the
    different reconstruction conditions, evaluated against the test views. Focusing
    first on the experiments using 50 input images, the baseline condition achieved
    by applying the original $R^{2}$-Gaussian framework \cite{zha_r2-gaussian_2024}
    to our synthetic circular dataset $\mathbf{I}_{\text{circ}}$, yielded the
    highest average scores with 39.0 dB PSNR and 0.97 SSIM. When using synthetic
    but arbitrarily posed $\mathbf{I}_{\text{DRR}}$ as input to our adapted
    framework, the performance decreased to 29.45 dB PSNR and 0.89 SSIM. The transition
    to using 50 arbitrarily posed real X-rays $\mathbf{I}_{\text{real}}$ resulted
    in a further decrease, yielding the lowest average PSNR of 23.22 dB and an
    SSIM of 0.76. Applying our proposed anatomy-guided radiographic standardization
    $\mathbf{I}_{\text{ST}}$ improved performance compared to using raw $\mathbf{I}
    _{\text{real}}$ inputs, reaching an average PSNR of 25.73 dB (an increase of
    2.51 dB) and an SSIM of 0.79, representing a partial recovery towards the $\mathbf{I}
    _{\text{DRR}}$ scores.

    The results obtained using 25 input images, also shown in Table~\ref{tab:results},
    followed similar trends but exhibited generally lower metric values across all
    conditions compared to the 50-image experiments. Notably, the relative improvement
    in PSNR when using $\mathbf{I}_{\text{ST}}$ compared to $\mathbf{I}_{\text{real}}$
    was more pronounced than the corresponding improvement in SSIM for both 50-view
    and 25-view cases.

    \label{sec:results:ablation}
    \begin{figure}
        \centering
        \includegraphics[width=0.6\linewidth]{
            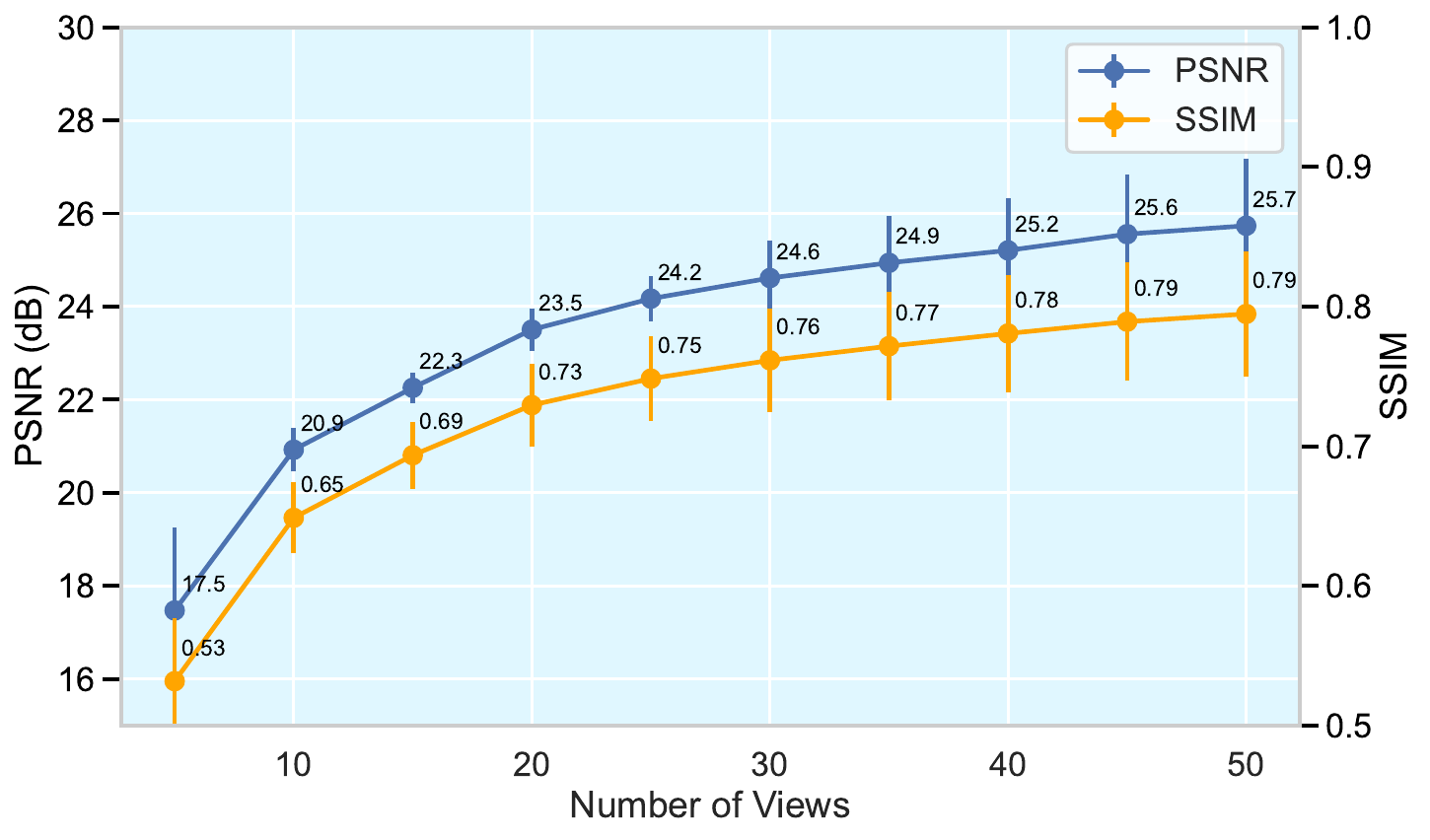
        }
        \caption{Evaluation of novel view synthesis quality from unseen poses
        using PSNR and SSIM metrics over varying numbers of views. Higher scores
        indicate better reconstruction quality.}
        \label{fig:gaussian:view_count}
    \end{figure}

    Figure~\ref{fig:gaussian:view_count} shows the PSNR/SSIM performance over a
    varying number of input $\mathbf{I}_{\text{ST}}$ views from 5 to 50 for one
    \textit{ex-vivo} specimen. The lowest average performance was observed with
    only five input images, yielding 17.5 dB PSNR and 0.53 SSIM. Both metrics
    showed a substantial increase up to 20 input images. Beyond 20 images, the
    rate of improvement decreased, with performance gradually reaching a maximum
    average of 25.7 dB PSNR and 0.79 SSIM at 50 input images. This quantitative
    trend, particularly the sharp initial improvement and later plateauing, can
    be compared with the qualitative usability assessments presented in Section~\ref{sec:results:qualitative_3d}.

    The original $R^{2}$-Gaussian paper \cite{zha_r2-gaussian_2024} reported
    performance for reconstructions from synthetic circular data, averaged over various
    objects, including two human torso CTs. They achieved average PSNR and SSIM values
    of 37.98 dB and 0.952, respectively, using 50 input images, and 35.19 dB and
    0.923, respectively, using 25 images. These results align well with our own baseline
    experiment using purely \textit{ex-vivo} synthetic circular data ($\mathbf{I}
    _{\text{circ}}$), where we achieved slightly higher values (39.0 dB / 0.97
    for 50 images). For their experiments on real, albeit circularly acquired, X-rays
    from the much simpler FIPS dataset \cite{noauthor_x-ray_nodate} consisting
    of objects like seashells and pine cones, they achieved 38.24 dB PSNR / 0.864
    SSIM for 50 images and 34.83 dB PSNR / 0.833 SSIM for 25 images. These
    scores, obtained under circular acquisition of simpler objects, are notably
    higher than those we obtained for our $\mathbf{I}_{\text{real}}$ condition, which
    involved reconstructions from real X-rays with complex anatomy and arbitrary
    poses.

    The average reconstruction time of IXGS using 50 $\mathbf{I}_{\text{ST}}$ images
    was 13m 33s on the hardware detailed in Section~\ref{sec:method:implementation}.
    For comparison, the original $R^{2}$-Gaussian paper \cite{zha_r2-gaussian_2024}
    also reported similar timings, such as 13m 52s for 30k training iterations. However,
    the authors of \cite{zha_r2-gaussian_2024} noted that high-quality results
    could be achieved much faster under specific conditions, reporting convergence
    in only 2m 35s on synthetic datasets using just 10k iterations.

\section*{Discussion}
    \label{sec:discussion} Our IXGS framework demonstrates that 3D volumetric reconstruction
    from sparse, arbitrarily posed intraoperative X-ray views using Gaussian
    splatting is feasible, using our proposed pipeline. While state-of-the-art
    methods benchmarked on synthetic data or real data from controlled circular
    trajectories show high fidelity \cite{zha_r2-gaussian_2024}, we validated our
    approach in a clinically relevant setting using real \textit{ex-vivo} data
    with sparse, arbitrary views mimicking intraoperative conditions. Crucially,
    our results indicate that clinically acceptable reconstructions ('Acceptable'
    to 'Very Good' surgeon ratings) can be achieved, particularly when
    approximately 20-30 or more views are available (Section~\ref{sec:results:qualitative_3d}),
    marking an important step towards practical surgical navigation applications.

    Our quantitative results (Section~\ref{sec:results:quantitative}) confirm a performance
    gap compared to idealized benchmarks. Reconstructions from arbitrary real X-rays
    ($\mathbf{I}_{\text{real}}$, $\mathbf{I}_{\text{ST}}$) yield lower PSNR/SSIM
    scores than those from synthetic circular ($\mathbf{I}_{\text{circ}}$) or even
    synthetic arbitrary ($\mathbf{I}_{\text{DRR}}$) data. Furthermore, early experiments
    indicated that simply increasing the number of Gaussians or tuning hyperparameters
    --- while potentially offering minor improvements for real data --- was insufficient
    to match the fidelity achieved with synthetic inputs and quickly became computationally
    infeasible. This highlights the combined difficulty posed by the domain gap
    (real vs. synthetic) and complex, arbitrary acquisition geometry. For instance,
    using 50 views, our $\mathbf{I}_{\text{ST}}$ reconstructions achieved an average
    PSNR of 25.73 dB, substantially lower than the 39.0 dB for
    $\mathbf{I}_{\text{circ}}$ but representing a necessary trade-off for
    clinical applicability.

    Despite modest quantitative scores (PSNR/SSIM), expert evaluation consistently
    indicated that the reconstructions, especially using
    $\mathbf{I}_{\text{ST}}$, provide anatomically useful representations for navigation
    (Section~\ref{sec:results:qualitative_3d}). This highlights the known
    limitations of standard metrics like PSNR and SSIM, which often correlate poorly
    with anatomical plausibility, diagnostic quality, or task-specific utility in
    complex medical image reconstruction scenarios \cite{breger_study_2024, dohmen_similarity_2025}.
    The ongoing development of learned perceptual metrics, such as LPIPS \cite{zhang_unreasonable_2018},
    indicates efforts to align evaluation with human visual assessment better. However,
    clinical task performance may require even more specific validation. Our observations
    reflect this gap: perceived reconstruction quality frequently improved
    beyond the saturation point of PSNR/SSIM during training or with added views.
    This supports the necessity of including qualitative, task-based assessments, like
    expert ratings, when evaluating clinical potential.

    The 3D evaluation further highlights the benefits of the proposed style transfer
    step ($\mathbf{I}_{\text{ST}}$ ). Notably, reconstructions generated directly
    from the original $\mathbf{I}_{\text{real}}$ images were found to be of insufficient
    quality for clinical usability assessment and were thus excluded from the expert
    rating process. This underscores the effectiveness of the
    $\mathbf{I}_{\text{ST}}$ approach, which improved the visibility of bone margins
    and overall anatomical clarity, leading to higher usability ratings (implicit
    from Fig.~\ref{fig:gaussian:comparison}, quantitative gain in Section~\ref{sec:results:quantitative},
    qualitative benefit in Section~\ref{sec:results:qualitative_3d}). This
    confirms that $\mathbf{I}_{\text{ST}}$ acts as an effective anatomy-guided
    radiographic standardization, improving consistency across views acquired under
    varying conditions and mitigating challenges from the domain gap and arbitrary
    geometry.

    Compared to previous work \cite{jecklin_domain_2024}, the proposed IXGS
    approach requires no anatomy-specific pretraining, enhancing its adaptability
    to new patients, pathologies, or anatomies. However, this flexibility comes at
    the cost of requiring more input views ($\sim$~20-30 for usability vs. 4 for
    X23D) and reconstruction time (13m 33s reported, Section~\ref{sec:results:quantitative}).
    While this time is significantly longer than the near-instantaneous
    inference (81.1ms) of a pre-trained model like our previous work X23D \cite{jecklin_domain_2024},
    convergence might potentially be accelerated using strategies like coarse
    initialization \cite{zha_r2-gaussian_2024}. Furthermore, the scalability to
    larger fields of view or higher resolutions without retraining remains an
    advantage over fixed-resolution methods.

    From a clinical perspective, achieving usable 3D volumes intraoperatively from
    sparse, arbitrary X-rays is highly relevant. The surgeon's feedback during the
    qualitative assessment provided noteworthy insights. Specifically, it was observed
    that the generated slice views were oriented parallel to the axes of the reconstruction
    volume itself, rather than necessarily representing true anatomical planes
    aligned perfectly with the patient's anatomy. This distinction was reportedly
    most noticeable in the axial views (e.g., Figure~\ref{fig:gaussian:slices}) and
    may have influenced usability ratings. Despite this observation about
    alignment, the surgeon also noted a general preference for the slice-based visualization
    over direct volume rendering, stating that scrolling through slices allowed
    for better spatial perception of anatomical relationships critical for
    navigation planning. This suggests that interactive slice viewers,
    potentially enhanced with improved anatomical alignment or artifact
    reduction, are a promising interface for navigation based on these reconstructions.
    The 'cloudy' artifacts observed outside the viewed regions (Fig.~\ref{fig:gaussian:comparison}),
    stemming from the Gaussian splatting initialization in unconstrained areas, are
    a limitation needing consideration, though often outside the primary
    surgical volume. Simple postprocessing or focusing on slice views can mitigate
    their impact.

    Current limitations include the reconstruction time (though potentially
    addressable with initialization strategies), the demonstrated dependency on
    view count and quality, characteristic reconstruction artifacts (e.g., 'cloudy'
    regions) in sparsely viewed areas, and visualization nuances such as volume-axis-aligned
    slices, which require consideration in downstream applications.

    Based on these findings, several future research directions emerge. Addressing
    robustness, particularly regarding challenging input views or outlier cases
    (like the poorly performing specimen in Section~\ref{sec:results:qualitative_3d}),
    is crucial. Improving visualization by mitigating artifacts or enabling
    anatomically aligned slicing requires investigation. Developing evaluation metrics
    that are better correlated with clinical utility remains an open challenge.
    Continued exploration of calibration-free workflows, coarse-to-fine reconstruction
    strategies for speed, and real-time updates integrated into navigation
    systems are promising avenues.

\section*{Conclusions}
    \label{sec:conclusions} We adapted Gaussian splatting within our proposed
    IXGS framework for 3D volumetric reconstruction from sparse ($\sim$20-30+
    views), arbitrarily posed real intraoperative X-rays, demonstrating feasibility
    beyond idealized circular or synthetic settings common in prior work. This
    extends neural rendering techniques toward clinically realistic C-arm
    imaging scenarios.

    Our proposed anatomy-guided radiographic standardization ($\mathbf{I}_{\text{ST}}$)
    using style transfer proved crucial. By reducing input appearance inconsistencies
    and enhancing relevant bone features, it simplified the reconstruction task
    for the Gaussian splatting optimization, yielding reconstructions with sufficient
    anatomical clarity for clinical navigation tasks, as confirmed by expert
    surgeon assessment.

    This clinical usability was achieved despite modest quantitative scores (PSNR/SSIM),
    highlighting the limitations of standard metrics for capturing task-specific
    utility. Furthermore, our approach requires no anatomy-specific pretraining,
    offering significant adaptability advantages over specialized deep learning
    models for diverse patients and pathologies.

    Future work should focus on accelerating reconstruction, improving robustness
    to challenging inputs and outlier cases, enhancing visualization methods to
    handle artifacts and potentially provide anatomically aligned slices, developing
    calibration-free approaches, establishing more clinically relevant
    evaluation metrics, and ultimately, integrating these techniques into real-time
    intraoperative navigation systems. This study represents a significant step
    towards leveraging neural rendering techniques for comprehensive 3D surgical
    guidance from standard C-arm fluoroscopy.

\bibliography{references}

\begin{thebibliography}{10}
\urlstyle{rm}
\expandafter\ifx\csname url\endcsname\relax
  \def\url#1{\texttt{#1}}\fi
\expandafter\ifx\csname urlprefix\endcsname\relax\def\urlprefix{URL }\fi
\expandafter\ifx\csname doiprefix\endcsname\relax\def\doiprefix{DOI: }\fi
\providecommand{\bibinfo}[2]{#2}
\providecommand{\eprint}[2][]{\url{#2}}

\bibitem{tonetti_role_2020}
\bibinfo{author}{Tonetti, J.}, \bibinfo{author}{Boudissa, M.}, \bibinfo{author}{Kerschbaumer, G.} \& \bibinfo{author}{Seurat, O.}
\newblock \bibinfo{journal}{\bibinfo{title}{Role of {3D} intraoperative imaging in orthopedic and trauma surgery}}.
\newblock {\emph{\JournalTitle{Orthopaedics \& Traumatology: Surgery \& Research}}} \textbf{\bibinfo{volume}{106}}, \bibinfo{pages}{S19--S25}, \doiprefix\url{10.1016/j.otsr.2019.05.021} (\bibinfo{year}{2020}).

\bibitem{keil_intraoperative_2023}
\bibinfo{author}{Keil, H.} \emph{et~al.}
\newblock \bibinfo{journal}{\bibinfo{title}{Intraoperative revision rates due to three-dimensional imaging in orthopedic trauma surgery: results of a case series of 4721 patients}}.
\newblock {\emph{\JournalTitle{European Journal of Trauma and Emergency Surgery: Official Publication of the European Trauma Society}}} \textbf{\bibinfo{volume}{49}}, \bibinfo{pages}{373--381}, \doiprefix\url{10.1007/s00068-022-02083-x} (\bibinfo{year}{2023}).

\bibitem{hong_effective_2010}
\bibinfo{author}{Hong, J.} \& \bibinfo{author}{Hashizume, M.}
\newblock \bibinfo{journal}{\bibinfo{title}{An effective point-based registration tool for surgical navigation}}.
\newblock {\emph{\JournalTitle{Surgical Endoscopy}}} \textbf{\bibinfo{volume}{24}}, \bibinfo{pages}{944--948}, \doiprefix\url{10.1007/s00464-009-0568-2} (\bibinfo{year}{2010}).

\bibitem{ma_autonomous_2020}
\bibinfo{author}{Ma, Q.} \emph{et~al.}
\newblock \bibinfo{journal}{\bibinfo{title}{Autonomous {Surgical} {Robot} {With} {Camera}-{Based} {Markerless} {Navigation} for {Oral} and {Maxillofacial} {Surgery}}}.
\newblock {\emph{\JournalTitle{IEEE/ASME Transactions on Mechatronics}}} \textbf{\bibinfo{volume}{25}}, \bibinfo{pages}{1084--1094}, \doiprefix\url{10.1109/TMECH.2020.2971618} (\bibinfo{year}{2020}).
\newblock \bibinfo{note}{Conference Name: IEEE/ASME Transactions on Mechatronics}.

\bibitem{suenaga_vision-based_2015}
\bibinfo{author}{Suenaga, H.} \emph{et~al.}
\newblock \bibinfo{journal}{\bibinfo{title}{Vision-based markerless registration using stereo vision and an augmented reality surgical navigation system: a pilot study}}.
\newblock {\emph{\JournalTitle{BMC Medical Imaging}}} \textbf{\bibinfo{volume}{15}}, \bibinfo{pages}{51}, \doiprefix\url{10.1186/s12880-015-0089-5} (\bibinfo{year}{2015}).

\bibitem{zheng_pairwise_2018}
\bibinfo{author}{Zheng, J.}, \bibinfo{author}{Miao, S.}, \bibinfo{author}{Wang, Z.~J.} \& \bibinfo{author}{Liao, R.}
\newblock \bibinfo{journal}{\bibinfo{title}{Pairwise domain adaptation module for {CNN}-based 2-{D}/3-{D} registration}}.
\newblock {\emph{\JournalTitle{Journal of Medical Imaging (Bellingham, Wash.)}}} \textbf{\bibinfo{volume}{5}}, \bibinfo{pages}{021204}, \doiprefix\url{10.1117/1.JMI.5.2.021204} (\bibinfo{year}{2018}).

\bibitem{ferrante_adaptability_2018}
\bibinfo{author}{Ferrante, E.}, \bibinfo{author}{Oktay, O.}, \bibinfo{author}{Glocker, B.} \& \bibinfo{author}{Milone, D.~H.}
\newblock \bibinfo{title}{On the {Adaptability} of {Unsupervised} {CNN}-{Based} {Deformable} {Image} {Registration} to {Unseen} {Image} {Domains}}.
\newblock In \bibinfo{editor}{Shi, Y.}, \bibinfo{editor}{Suk, H.-I.} \& \bibinfo{editor}{Liu, M.} (eds.) \emph{\bibinfo{booktitle}{Machine {Learning} in {Medical} {Imaging}}}, \bibinfo{pages}{294--302}, \doiprefix\url{10.1007/978-3-030-00919-9_34} (\bibinfo{publisher}{Springer International Publishing}, \bibinfo{address}{Cham}, \bibinfo{year}{2018}).

\bibitem{zhang_risk_2019}
\bibinfo{author}{Zhang, J.~N.}, \bibinfo{author}{Fan, Y.} \& \bibinfo{author}{Hao, D.~J.}
\newblock \bibinfo{journal}{\bibinfo{title}{Risk factors for robot-assisted spinal pedicle screw malposition}}.
\newblock {\emph{\JournalTitle{Scientific Reports}}} \textbf{\bibinfo{volume}{9}}, \bibinfo{pages}{3025}, \doiprefix\url{10.1038/s41598-019-40057-z} (\bibinfo{year}{2019}).
\newblock \bibinfo{note}{Publisher: Nature Publishing Group}.

\bibitem{rahmathulla_intraoperative_2014}
\bibinfo{author}{Rahmathulla, G.}, \bibinfo{author}{Nottmeier, E.~W.}, \bibinfo{author}{Pirris, S.~M.}, \bibinfo{author}{Deen, H.~G.} \& \bibinfo{author}{Pichelmann, M.~A.}
\newblock \bibinfo{journal}{\bibinfo{title}{Intraoperative image-guided spinal navigation: technical pitfalls and their avoidance}}.
\newblock {\emph{\JournalTitle{Neurosurgical Focus}}} \textbf{\bibinfo{volume}{36}}, \bibinfo{pages}{E3}, \doiprefix\url{10.3171/2014.1.FOCUS13516} (\bibinfo{year}{2014}).

\bibitem{venkatesh_cone_2017}
\bibinfo{author}{Venkatesh, E.} \& \bibinfo{author}{Elluru, S.~V.}
\newblock \bibinfo{journal}{\bibinfo{title}{Cone beam computed tomography: basics and applications in dentistry}}.
\newblock {\emph{\JournalTitle{Journal of Istanbul University Faculty of Dentistry}}} \textbf{\bibinfo{volume}{51}}, \bibinfo{pages}{S102--S121}, \doiprefix\url{10.17096/jiufd.00289} (\bibinfo{year}{2017}).

\bibitem{costa_spinal_2011}
\bibinfo{author}{Costa, F.} \emph{et~al.}
\newblock \bibinfo{journal}{\bibinfo{title}{Spinal {Navigation}: {Standard} {Preoperative}: {Versus}: {Intraoperative} {Computed} {Tomography} {Data} {Set} {Acquisition} for {Computer}-{Guidance} {System}: {Radiological} and {Clinical} {Study} in 100 {Consecutive} {Patients}}}.
\newblock {\emph{\JournalTitle{Spine}}} \textbf{\bibinfo{volume}{36}}, \bibinfo{pages}{2094--2098}, \doiprefix\url{10.1097/BRS.0b013e318201129d} (\bibinfo{year}{2011}).

\bibitem{mendelsohn_patient_2016}
\bibinfo{author}{Mendelsohn, D.} \emph{et~al.}
\newblock \bibinfo{journal}{\bibinfo{title}{Patient and surgeon radiation exposure during spinal instrumentation using intraoperative computed tomography-based navigation}}.
\newblock {\emph{\JournalTitle{The Spine Journal}}} \textbf{\bibinfo{volume}{16}}, \bibinfo{pages}{343--354}, \doiprefix\url{10.1016/j.spinee.2015.11.020} (\bibinfo{year}{2016}).

\bibitem{villard_radiation_2014}
\bibinfo{author}{Villard, J.} \emph{et~al.}
\newblock \bibinfo{journal}{\bibinfo{title}{Radiation {Exposure} to the {Surgeon} and the {Patient} {During} {Posterior} {Lumbar} {Spinal} {Instrumentation}: {A} {Prospective} {Randomized} {Comparison} of {Navigated} {Versus} {Non}-navigated {Freehand} {Techniques}}}.
\newblock {\emph{\JournalTitle{Spine}}} \textbf{\bibinfo{volume}{39}}, \bibinfo{pages}{1004}, \doiprefix\url{10.1097/BRS.0000000000000351} (\bibinfo{year}{2014}).

\bibitem{dea_economic_2016}
\bibinfo{author}{Dea, N.} \emph{et~al.}
\newblock \bibinfo{journal}{\bibinfo{title}{Economic evaluation comparing intraoperative cone beam {CT}-based navigation and conventional fluoroscopy for the placement of spinal pedicle screws: a patient-level data cost-effectiveness analysis}}.
\newblock {\emph{\JournalTitle{The Spine Journal}}} \textbf{\bibinfo{volume}{16}}, \bibinfo{pages}{23--31}, \doiprefix\url{10.1016/j.spinee.2015.09.062} (\bibinfo{year}{2016}).

\bibitem{beck_benefit_2009}
\bibinfo{author}{Beck, M.}, \bibinfo{author}{Mittlmeier, T.}, \bibinfo{author}{Gierer, P.}, \bibinfo{author}{Harms, C.} \& \bibinfo{author}{Gradl, G.}
\newblock \bibinfo{journal}{\bibinfo{title}{Benefit and accuracy of intraoperative {3D}-imaging after pedicle screw placement: a prospective study in stabilizing thoracolumbar fractures}}.
\newblock {\emph{\JournalTitle{European Spine Journal}}} \textbf{\bibinfo{volume}{18}}, \bibinfo{pages}{1469--1477}, \doiprefix\url{10.1007/s00586-009-1050-5} (\bibinfo{year}{2009}).

\bibitem{jecklin_x23dintraoperative_2022}
\bibinfo{author}{Jecklin, S.}, \bibinfo{author}{Jancik, C.}, \bibinfo{author}{Farshad, M.}, \bibinfo{author}{Fürnstahl, P.} \& \bibinfo{author}{Esfandiari, H.}
\newblock \bibinfo{journal}{\bibinfo{title}{{X23D}—intraoperative {3D} lumbar spine shape reconstruction based on sparse multi-view {X}-ray data}}.
\newblock {\emph{\JournalTitle{Journal of Imaging}}} \textbf{\bibinfo{volume}{8}}, \bibinfo{pages}{271}, \doiprefix\url{10.3390/jimaging8100271} (\bibinfo{year}{2022}).
\newblock \bibinfo{note}{Publisher: MDPI}.

\bibitem{jecklin_domain_2024}
\bibinfo{author}{Jecklin, S.} \emph{et~al.}
\newblock \bibinfo{journal}{\bibinfo{title}{Domain adaptation strategies for {3D} reconstruction of the lumbar spine using real fluoroscopy data}}.
\newblock {\emph{\JournalTitle{Medical Image Analysis}}} \textbf{\bibinfo{volume}{98}}, \bibinfo{pages}{103322}, \doiprefix\url{10.1016/j.media.2024.103322} (\bibinfo{year}{2024}).
\newblock \bibinfo{note}{Publisher: Elsevier}.

\bibitem{zha_r2-gaussian_2024}
\bibinfo{author}{Zha, R.} \emph{et~al.}
\newblock \bibinfo{journal}{\bibinfo{title}{R\${\textasciicircum}2\$-{Gaussian}: {Rectifying} {Radiative} {Gaussian} {Splatting} for {Tomographic} {Reconstruction}}}.
\newblock {\emph{\JournalTitle{Advances in Neural Information Processing Systems}}} \textbf{\bibinfo{volume}{37}}, \bibinfo{pages}{44907--44934} (\bibinfo{year}{2024}).

\bibitem{zha_naf_2022}
\bibinfo{author}{Zha, R.}, \bibinfo{author}{Zhang, Y.} \& \bibinfo{author}{Li, H.}
\newblock \bibinfo{title}{{NAF}: {Neural} {Attenuation} {Fields} for {Sparse}-{View} {CBCT} {Reconstruction}}.
\newblock In \bibinfo{editor}{Wang, L.}, \bibinfo{editor}{Dou, Q.}, \bibinfo{editor}{Fletcher, P.~T.}, \bibinfo{editor}{Speidel, S.} \& \bibinfo{editor}{Li, S.} (eds.) \emph{\bibinfo{booktitle}{Medical {Image} {Computing} and {Computer} {Assisted} {Intervention} – {MICCAI} 2022}}, \bibinfo{pages}{442--452}, \doiprefix\url{10.1007/978-3-031-16446-0_42} (\bibinfo{publisher}{Springer Nature Switzerland}, \bibinfo{address}{Cham}, \bibinfo{year}{2022}).

\bibitem{ruckert_neat_2022}
\bibinfo{author}{Rückert, D.}, \bibinfo{author}{Wang, Y.}, \bibinfo{author}{Li, R.}, \bibinfo{author}{Idoughi, R.} \& \bibinfo{author}{Heidrich, W.}
\newblock \bibinfo{journal}{\bibinfo{title}{{NeAT}: neural adaptive tomography}}.
\newblock {\emph{\JournalTitle{ACM Trans. Graph.}}} \textbf{\bibinfo{volume}{41}}, \bibinfo{pages}{55:1--55:13}, \doiprefix\url{10.1145/3528223.3530121} (\bibinfo{year}{2022}).

\bibitem{rangelov_impact_2024}
\bibinfo{author}{Rangelov, D.}, \bibinfo{author}{Waanders, S.}, \bibinfo{author}{Waanders, K.}, \bibinfo{author}{van Keulen, M.} \& \bibinfo{author}{Miltchev, R.}
\newblock \bibinfo{journal}{\bibinfo{title}{Impact of {Camera} {Settings} on {3D} {Reconstruction} {Quality}: {Insights} from {NeRF} and {Gaussian} {Splatting}}}.
\newblock {\emph{\JournalTitle{Sensors (Basel, Switzerland)}}} \textbf{\bibinfo{volume}{24}}, \bibinfo{pages}{7594}, \doiprefix\url{10.3390/s24237594} (\bibinfo{year}{2024}).

\bibitem{ye_gaussian_2024}
\bibinfo{author}{Ye, S.}, \bibinfo{author}{Dong, Z.}, \bibinfo{author}{Hu, Y.}, \bibinfo{author}{Wen, Y.} \& \bibinfo{author}{Liu, Y.}
\newblock \bibinfo{journal}{\bibinfo{title}{Gaussian in the {Dark}: {Real}‐{Time} {View} {Synthesis} {From} {Inconsistent} {Dark} {Images} {Using} {Gaussian} {Splatting}}}.
\newblock {\emph{\JournalTitle{Computer Graphics Forum}}} \textbf{\bibinfo{volume}{43}}, \bibinfo{pages}{e15213}, \doiprefix\url{10.1111/cgf.15213} (\bibinfo{year}{2024}).
\newblock \bibinfo{note}{\_eprint: https://onlinelibrary.wiley.com/doi/pdf/10.1111/cgf.15213}.

\bibitem{isola_image--image_2018}
\bibinfo{author}{Isola, P.}, \bibinfo{author}{Zhu, J.-Y.}, \bibinfo{author}{Zhou, T.} \& \bibinfo{author}{Efros, A.~A.}
\newblock \bibinfo{title}{Image-to-{Image} {Translation} with {Conditional} {Adversarial} {Networks}}, \doiprefix\url{10.48550/arXiv.1611.07004} (\bibinfo{year}{2018}).
\newblock \bibinfo{note}{ArXiv:1611.07004 [cs]}.

\bibitem{kasten_end--end_2020}
\bibinfo{author}{Kasten, Y.}, \bibinfo{author}{Doktofsky, D.} \& \bibinfo{author}{Kovler, I.}
\newblock \bibinfo{title}{End-{To}-{End} {Convolutional} {Neural} {Network} for {3D} {Reconstruction} of {Knee} {Bones} from {Bi}-planar {X}-{Ray} {Images}}.
\newblock In \bibinfo{editor}{Deeba, F.}, \bibinfo{editor}{Johnson, P.}, \bibinfo{editor}{Würfl, T.} \& \bibinfo{editor}{Ye, J.~C.} (eds.) \emph{\bibinfo{booktitle}{Machine {Learning} for {Medical} {Image} {Reconstruction}}}, \bibinfo{pages}{123--133}, \doiprefix\url{10.1007/978-3-030-61598-7_12} (\bibinfo{publisher}{Springer International Publishing}, \bibinfo{address}{Cham}, \bibinfo{year}{2020}).

\bibitem{shiode_2d3d_2021}
\bibinfo{author}{Shiode, R.} \emph{et~al.}
\newblock \bibinfo{journal}{\bibinfo{title}{{2D}–{3D} reconstruction of distal forearm bone from actual {X}-ray images of the wrist using convolutional neural networks}}.
\newblock {\emph{\JournalTitle{Scientific Reports}}} \textbf{\bibinfo{volume}{11}}, \bibinfo{pages}{15249}, \doiprefix\url{10.1038/s41598-021-94634-2} (\bibinfo{year}{2021}).
\newblock \bibinfo{note}{Publisher: Nature Publishing Group}.

\bibitem{ge_x-ctrsnet_2022}
\bibinfo{author}{Ge, R.} \emph{et~al.}
\newblock \bibinfo{journal}{\bibinfo{title}{X-{CTRSNet}: {3D} cervical vertebra {CT} reconstruction and segmentation directly from {2D} {X}-ray images}}.
\newblock {\emph{\JournalTitle{Knowledge-Based Systems}}} \textbf{\bibinfo{volume}{236}}, \bibinfo{pages}{107680}, \doiprefix\url{10.1016/j.knosys.2021.107680} (\bibinfo{year}{2022}).

\bibitem{kar_learning_2017}
\bibinfo{author}{Kar, A.}, \bibinfo{author}{Häne, C.} \& \bibinfo{author}{Malik, J.}
\newblock \bibinfo{journal}{\bibinfo{title}{Learning a {Multi}-{View} {Stereo} {Machine}}}.
\newblock {\emph{\JournalTitle{arXiv:1708.05375 [cs]}}}  (\bibinfo{year}{2017}).

\bibitem{luchmann_spinal_2024}
\bibinfo{author}{Luchmann, D.} \emph{et~al.}
\newblock \bibinfo{journal}{\bibinfo{title}{Spinal navigation with {AI}-driven {3D}-reconstruction of fluoroscopy images: an ex-vivo feasibility study}}.
\newblock {\emph{\JournalTitle{BMC Musculoskeletal Disorders}}} \textbf{\bibinfo{volume}{25}}, \bibinfo{pages}{925}, \doiprefix\url{10.1186/s12891-024-08052-2} (\bibinfo{year}{2024}).
\newblock \bibinfo{note}{Publisher: BioMed Central London}.

\bibitem{mildenhall_nerf_2021}
\bibinfo{author}{Mildenhall, B.} \emph{et~al.}
\newblock \bibinfo{journal}{\bibinfo{title}{{NeRF}: representing scenes as neural radiance fields for view synthesis}}.
\newblock {\emph{\JournalTitle{Commun. ACM}}} \textbf{\bibinfo{volume}{65}}, \bibinfo{pages}{99--106}, \doiprefix\url{10.1145/3503250} (\bibinfo{year}{2021}).

\bibitem{kerbl_3d_2023}
\bibinfo{author}{Kerbl, B.}, \bibinfo{author}{Kopanas, G.}, \bibinfo{author}{Leimkuehler, T.} \& \bibinfo{author}{Drettakis, G.}
\newblock \bibinfo{journal}{\bibinfo{title}{{3D} {Gaussian} {Splatting} for {Real}-{Time} {Radiance} {Field} {Rendering}}}.
\newblock {\emph{\JournalTitle{ACM Trans. Graph.}}} \textbf{\bibinfo{volume}{42}}, \bibinfo{pages}{139:1--139:14}, \doiprefix\url{10.1145/3592433} (\bibinfo{year}{2023}).

\bibitem{rakotosaona_nerfmeshing_2024}
\bibinfo{author}{Rakotosaona, M.-J.} \emph{et~al.}
\newblock \bibinfo{title}{{NeRFMeshing}: {Distilling} {Neural} {Radiance} {Fields} into {Geometrically}-{Accurate} {3D} {Meshes}}.
\newblock In \emph{\bibinfo{booktitle}{2024 {International} {Conference} on {3D} {Vision} ({3DV})}}, \bibinfo{pages}{1156--1165}, \doiprefix\url{10.1109/3DV62453.2024.00093} (\bibinfo{year}{2024}).
\newblock \bibinfo{note}{ISSN: 2475-7888}.

\bibitem{corona-figueroa_mednerf_2022}
\bibinfo{author}{Corona-Figueroa, A.} \emph{et~al.}
\newblock \bibinfo{journal}{\bibinfo{title}{{MedNeRF}: {Medical} {Neural} {Radiance} {Fields} for {Reconstructing} {3D}-aware {CT}-{Projections} from a {Single} {X}-ray}}.
\newblock {\emph{\JournalTitle{2022 44th Annual International Conference of the IEEE Engineering in Medicine \& Biology Society (EMBC)}}} \bibinfo{pages}{3843--3848}, \doiprefix\url{10.1109/EMBC48229.2022.9871757} (\bibinfo{year}{2022}).
\newblock \bibinfo{note}{Conference Name: 2022 44th Annual International Conference of the IEEE Engineering in Medicine \& Biology Society (EMBC) ISBN: 9781728127828 Place: Glasgow, Scotland, United Kingdom Publisher: IEEE}.

\bibitem{cai_radiative_2025}
\bibinfo{author}{Cai, Y.} \emph{et~al.}
\newblock \bibinfo{title}{Radiative {Gaussian} {Splatting} for {Efficient} {X}-{Ray} {Novel} {View} {Synthesis}}.
\newblock In \bibinfo{editor}{Leonardis, A.} \emph{et~al.} (eds.) \emph{\bibinfo{booktitle}{Computer {Vision} – {ECCV} 2024}}, \bibinfo{pages}{283--299}, \doiprefix\url{10.1007/978-3-031-73232-4_16} (\bibinfo{publisher}{Springer Nature Switzerland}, \bibinfo{address}{Cham}, \bibinfo{year}{2025}).

\bibitem{gao_ddgs-ct_2024}
\bibinfo{author}{Gao, Z.} \emph{et~al.}
\newblock \bibinfo{journal}{\bibinfo{title}{{DDGS}-{CT}: {Direction}-{Disentangled} {Gaussian} {Splatting} for {Realistic} {Volume} {Rendering}}}.
\newblock {\emph{\JournalTitle{Advances in Neural Information Processing Systems}}} \textbf{\bibinfo{volume}{37}}, \bibinfo{pages}{39281--39302} (\bibinfo{year}{2024}).

\bibitem{wysocki_ultra-nerf_2024}
\bibinfo{author}{Wysocki, M.} \emph{et~al.}
\newblock \bibinfo{title}{Ultra-{NeRF}: {Neural} {Radiance} {Fields} for {Ultrasound} {Imaging}}.
\newblock In \emph{\bibinfo{booktitle}{Medical {Imaging} with {Deep} {Learning}}}, \bibinfo{pages}{382--401} (\bibinfo{publisher}{PMLR}, \bibinfo{year}{2024}).
\newblock \bibinfo{note}{ISSN: 2640-3498}.

\bibitem{awojoyogbe_neural_2024}
\bibinfo{author}{Awojoyogbe, B.~O.} \& \bibinfo{author}{Dada, M.~O.}
\newblock \bibinfo{title}{Neural {Radiance} {Fields} ({NeRFs}) {Technique} to {Render} {3D} {Reconstruction} of {Magnetic} {Resonance} {Images}}.
\newblock In \bibinfo{editor}{Awojoyogbe, B.~O.} \& \bibinfo{editor}{Dada, M.~O.} (eds.) \emph{\bibinfo{booktitle}{Digital {Molecular} {Magnetic} {Resonance} {Imaging}}}, \bibinfo{pages}{247--258}, \doiprefix\url{10.1007/978-981-97-6370-2_10} (\bibinfo{publisher}{Springer Nature}, \bibinfo{address}{Singapore}, \bibinfo{year}{2024}).

\bibitem{yang_deform3dgs_2024}
\bibinfo{author}{Yang, S.} \emph{et~al.}
\newblock \bibinfo{title}{{Deform3DGS}: {Flexible} {Deformation} for {Fast} {Surgical} {Scene} {Reconstruction} with {Gaussian} {Splatting}}.
\newblock In \bibinfo{editor}{Linguraru, M.~G.} \emph{et~al.} (eds.) \emph{\bibinfo{booktitle}{Medical {Image} {Computing} and {Computer} {Assisted} {Intervention} – {MICCAI} 2024}}, \bibinfo{pages}{132--142}, \doiprefix\url{10.1007/978-3-031-72089-5_13} (\bibinfo{publisher}{Springer Nature Switzerland}, \bibinfo{address}{Cham}, \bibinfo{year}{2024}).

\bibitem{cai_structure-aware_2024}
\bibinfo{author}{Cai, Y.}, \bibinfo{author}{Wang, J.}, \bibinfo{author}{Yuille, A.}, \bibinfo{author}{Zhou, Z.} \& \bibinfo{author}{Wang, A.}
\newblock \bibinfo{title}{Structure-{Aware} {Sparse}-{View} {X}-ray {3D} {Reconstruction}}.
\newblock \bibinfo{pages}{11174--11183} (\bibinfo{year}{2024}).

\bibitem{noauthor_x-ray_nodate}
\bibinfo{title}{X-ray {Tomographic} {Datasets} {Archives}}.

\bibitem{chen_vertxnet_2024}
\bibinfo{author}{Chen, Y.} \emph{et~al.}
\newblock \bibinfo{journal}{\bibinfo{title}{{VertXNet}: an ensemble method for vertebral body segmentation and identification from cervical and lumbar spinal {X}-rays}}.
\newblock {\emph{\JournalTitle{Scientific Reports}}} \textbf{\bibinfo{volume}{14}}, \bibinfo{pages}{3341}, \doiprefix\url{10.1038/s41598-023-49923-3} (\bibinfo{year}{2024}).
\newblock \bibinfo{note}{Publisher: Nature Publishing Group}.

\bibitem{kim_automatic_2021}
\bibinfo{author}{Kim, K.~C.}, \bibinfo{author}{Cho, H.~C.}, \bibinfo{author}{Jang, T.~J.}, \bibinfo{author}{Choi, J.~M.} \& \bibinfo{author}{Seo, J.~K.}
\newblock \bibinfo{journal}{\bibinfo{title}{Automatic detection and segmentation of lumbar vertebrae from {X}-ray images for compression fracture evaluation}}.
\newblock {\emph{\JournalTitle{Computer Methods and Programs in Biomedicine}}} \textbf{\bibinfo{volume}{200}}, \bibinfo{pages}{105833}, \doiprefix\url{10.1016/j.cmpb.2020.105833} (\bibinfo{year}{2021}).

\bibitem{darcet_vision_2024}
\bibinfo{author}{Darcet, T.}, \bibinfo{author}{Oquab, M.}, \bibinfo{author}{Mairal, J.} \& \bibinfo{author}{Bojanowski, P.}
\newblock \bibinfo{title}{Vision {Transformers} {Need} {Registers}}, \doiprefix\url{10.48550/arXiv.2309.16588} (\bibinfo{year}{2024}).
\newblock \bibinfo{note}{ArXiv:2309.16588 [cs]}.

\bibitem{kirillov_segment_2023}
\bibinfo{author}{Kirillov, A.} \emph{et~al.}
\newblock \bibinfo{title}{Segment {Anything}}.
\newblock \bibinfo{pages}{4015--4026} (\bibinfo{year}{2023}).

\bibitem{wu_medical_2023}
\bibinfo{author}{Wu, J.} \emph{et~al.}
\newblock \bibinfo{title}{Medical {SAM} {Adapter}: {Adapting} {Segment} {Anything} {Model} for {Medical} {Image} {Segmentation}}, \doiprefix\url{10.48550/arXiv.2304.12620} (\bibinfo{year}{2023}).
\newblock \bibinfo{note}{ArXiv:2304.12620 [cs]}.

\bibitem{ehab_unet_2024}
\bibinfo{author}{Ehab, W.}, \bibinfo{author}{Huang, L.} \& \bibinfo{author}{Li, Y.}
\newblock \bibinfo{journal}{\bibinfo{title}{{UNet} and {Variants} for {Medical} {Image} {Segmentation}}}.
\newblock {\emph{\JournalTitle{International Journal of Network Dynamics and Intelligence}}} \bibinfo{pages}{100009--100009}, \doiprefix\url{10.53941/ijndi.2024.100009} (\bibinfo{year}{2024}).

\bibitem{gupta_vitol_2022}
\bibinfo{author}{Gupta, S.}, \bibinfo{author}{Lakhotia, S.}, \bibinfo{author}{Rawat, A.} \& \bibinfo{author}{Tallamraju, R.}
\newblock \bibinfo{title}{{ViTOL}: {Vision} {Transformer} for {Weakly} {Supervised} {Object} {Localization}}.
\newblock \bibinfo{pages}{4101--4110} (\bibinfo{year}{2022}).

\bibitem{yookwan_coarse_2023}
\bibinfo{author}{Yookwan, W.} \emph{et~al.}
\newblock \bibinfo{journal}{\bibinfo{title}{Coarse {X}-ray {Lumbar} {Vertebrae} {Pose} {Localization} and {Registration} {Using} {Triangulation} {Correspondence}}}.
\newblock {\emph{\JournalTitle{Processes}}} \textbf{\bibinfo{volume}{11}}, \bibinfo{pages}{61}, \doiprefix\url{10.3390/pr11010061} (\bibinfo{year}{2023}).
\newblock \bibinfo{note}{Number: 1 Publisher: Multidisciplinary Digital Publishing Institute}.

\bibitem{ye_projective-geometry-aware_2025}
\bibinfo{author}{Ye, K.}, \bibinfo{author}{Sun, W.}, \bibinfo{author}{Tao, R.} \& \bibinfo{author}{Zheng, G.}
\newblock \bibinfo{journal}{\bibinfo{title}{A {Projective}-{Geometry}-{Aware} {Network} for {3D} {Vertebra} {Localization} in {Calibrated} {Biplanar} {X}-{Ray} {Images}}}.
\newblock {\emph{\JournalTitle{Sensors}}} \textbf{\bibinfo{volume}{25}}, \bibinfo{pages}{1123}, \doiprefix\url{10.3390/s25041123} (\bibinfo{year}{2025}).
\newblock \bibinfo{note}{Number: 4 Publisher: Multidisciplinary Digital Publishing Institute}.

\bibitem{wang_yolov7_2023}
\bibinfo{author}{Wang, C.-Y.}, \bibinfo{author}{Bochkovskiy, A.} \& \bibinfo{author}{Liao, H.-Y.~M.}
\newblock \bibinfo{title}{{YOLOv7}: {Trainable} {Bag}-of-{Freebies} {Sets} {New} {State}-of-the-{Art} for {Real}-{Time} {Object} {Detectors}}.
\newblock \bibinfo{pages}{7464--7475} (\bibinfo{year}{2023}).

\bibitem{zhu_unpaired_2017}
\bibinfo{author}{Zhu, J.-Y.}, \bibinfo{author}{Park, T.}, \bibinfo{author}{Isola, P.} \& \bibinfo{author}{Efros, A.~A.}
\newblock \bibinfo{title}{Unpaired {Image}-{To}-{Image} {Translation} {Using} {Cycle}-{Consistent} {Adversarial} {Networks}}.
\newblock In \emph{\bibinfo{booktitle}{2017 {IEEE} {International} {Conference} on {Computer} {Vision} ({ICCV})}}, \bibinfo{pages}{2223--2232}, \doiprefix\url{10.1109/ICCV.2017.244} (\bibinfo{publisher}{IEEE}, \bibinfo{address}{Venice}, \bibinfo{year}{2017}).

\bibitem{duda_use_1972}
\bibinfo{author}{Duda, R.~O.} \& \bibinfo{author}{Hart, P.~E.}
\newblock \bibinfo{journal}{\bibinfo{title}{Use of the {Hough} transformation to detect lines and curves in pictures}}.
\newblock {\emph{\JournalTitle{Commun. ACM}}} \textbf{\bibinfo{volume}{15}}, \bibinfo{pages}{11--15}, \doiprefix\url{10.1145/361237.361242} (\bibinfo{year}{1972}).

\bibitem{abdel-aziz_direct_2015}
\bibinfo{author}{Abdel-Aziz, Y.~I.}, \bibinfo{author}{Karara, H.~M.} \& \bibinfo{author}{Hauck, M.}
\newblock \bibinfo{journal}{\bibinfo{title}{Direct {Linear} {Transformation} from {Comparator} {Coordinates} into {Object} {Space} {Coordinates} in {Close}-{Range} {Photogrammetry}*}}.
\newblock {\emph{\JournalTitle{Photogrammetric Engineering \& Remote Sensing}}} \textbf{\bibinfo{volume}{81}}, \bibinfo{pages}{103--107}, \doiprefix\url{10.14358/PERS.81.2.103} (\bibinfo{year}{2015}).

\bibitem{feldkamp_practical_1984}
\bibinfo{author}{Feldkamp, L.~A.}, \bibinfo{author}{Davis, L.~C.} \& \bibinfo{author}{Kress, J.~W.}
\newblock \bibinfo{journal}{\bibinfo{title}{Practical cone-beam algorithm}}.
\newblock {\emph{\JournalTitle{JOSA A}}} \textbf{\bibinfo{volume}{1}}, \bibinfo{pages}{612--619}, \doiprefix\url{10.1364/JOSAA.1.000612} (\bibinfo{year}{1984}).
\newblock \bibinfo{note}{Publisher: Optica Publishing Group}.

\bibitem{zwicker_ewa_2002}
\bibinfo{author}{Zwicker, M.}, \bibinfo{author}{Pfister, H.}, \bibinfo{author}{van Baar, J.} \& \bibinfo{author}{Gross, M.}
\newblock \bibinfo{journal}{\bibinfo{title}{{EWA} splatting}}.
\newblock {\emph{\JournalTitle{IEEE Transactions on Visualization and Computer Graphics}}} \textbf{\bibinfo{volume}{8}}, \bibinfo{pages}{223--238}, \doiprefix\url{10.1109/TVCG.2002.1021576} (\bibinfo{year}{2002}).

\bibitem{wang_image_2004}
\bibinfo{author}{Wang, Z.}, \bibinfo{author}{Bovik, A.}, \bibinfo{author}{Sheikh, H.} \& \bibinfo{author}{Simoncelli, E.}
\newblock \bibinfo{journal}{\bibinfo{title}{Image quality assessment: from error visibility to structural similarity}}.
\newblock {\emph{\JournalTitle{IEEE Transactions on Image Processing}}} \textbf{\bibinfo{volume}{13}}, \bibinfo{pages}{600--612}, \doiprefix\url{10.1109/TIP.2003.819861} (\bibinfo{year}{2004}).
\newblock \bibinfo{note}{Conference Name: IEEE Transactions on Image Processing}.

\bibitem{rudin_nonlinear_1992}
\bibinfo{author}{Rudin, L.~I.}, \bibinfo{author}{Osher, S.} \& \bibinfo{author}{Fatemi, E.}
\newblock \bibinfo{journal}{\bibinfo{title}{Nonlinear total variation based noise removal algorithms}}.
\newblock {\emph{\JournalTitle{Physica D: Nonlinear Phenomena}}} \textbf{\bibinfo{volume}{60}}, \bibinfo{pages}{259--268}, \doiprefix\url{10.1016/0167-2789(92)90242-F} (\bibinfo{year}{1992}).

\bibitem{breger_study_2024}
\bibinfo{author}{Breger, A.} \emph{et~al.}
\newblock \bibinfo{title}{A study on the adequacy of common {IQA} measures for medical images}, \doiprefix\url{10.48550/arXiv.2405.19224} (\bibinfo{year}{2024}).
\newblock \bibinfo{note}{ArXiv:2405.19224 [eess]}.

\bibitem{dohmen_similarity_2025}
\bibinfo{author}{Dohmen, M.}, \bibinfo{author}{Klemens, M.~A.}, \bibinfo{author}{Baltruschat, I.~M.}, \bibinfo{author}{Truong, T.} \& \bibinfo{author}{Lenga, M.}
\newblock \bibinfo{journal}{\bibinfo{title}{Similarity and quality metrics for {MR} image-to-image translation}}.
\newblock {\emph{\JournalTitle{Scientific Reports}}} \textbf{\bibinfo{volume}{15}}, \bibinfo{pages}{3853}, \doiprefix\url{10.1038/s41598-025-87358-0} (\bibinfo{year}{2025}).
\newblock \bibinfo{note}{Publisher: Nature Publishing Group}.

\bibitem{zhang_unreasonable_2018}
\bibinfo{author}{Zhang, R.}, \bibinfo{author}{Isola, P.}, \bibinfo{author}{Efros, A.~A.}, \bibinfo{author}{Shechtman, E.} \& \bibinfo{author}{Wang, O.}
\newblock \bibinfo{title}{The {Unreasonable} {Effectiveness} of {Deep} {Features} as a {Perceptual} {Metric}}.
\newblock \bibinfo{pages}{586--595} (\bibinfo{year}{2018}).

\end{thebibliography}

\section*{Acknowledgements} 
This work has been supported by the OR-X - a Swiss national research infrastructure for translational surgery and associated funding by the University Hospital Balgrist . We like to thank Tanja Walther for her support during the data capture. Financial support was provided by the Monique Dornonville de la Cour Foundation and an internal Balgrist University Hospital fund.

\section*{Author contributions statement}
S.J.: Conceptualization, Methodology, Software, Validation, Formal analysis, Investigation, Data Curation, Writing - Original Draft, Visualization. 
A.M.: Software, Writing - Review \& Editing. 
R.Z.: Software, Methodology, Writing - Review \& Editing.
L.C.: Software, Validation.
C.J.L.: Investigation (Clinical Evaluation), Validation.
M.F.: Resources, Funding acquisition. 
P.F.: Conceptualization, Resources, Writing - Review \& Editing, Supervision, Funding acquisition, Project administration. 
All authors reviewed the manuscript.

\section*{Data Availability}
The code adapted for this study, along with a small dataset for testing derived from the \textit{ex-vivo} data, will be made available at \url{https://github.com/MrMonk3y/IXGS}. The primary \textit{ex-vivo} paired dataset (real X-rays, DRRs, calibration data) generated and analysed during the current study is not publicly available due to restrictions relating to pending patent applications but is available from the corresponding author on reasonable request.

\section*{Competing Interests} 
S.J. reports that financial support for his doctoral studies was provided by the Monique Dornonville de la Cour Foundation and an internal Balgrist University Hospital fund. 
M.F. reports a relationship with X23D AG that includes equity or stocks. 
P.F. reports a relationship with X23D AG that includes board membership and equity or stocks.
P.F. and M.F. have patent $\#$WO2023156608A1 pending to University of Zurich related to prior work. 
P.F., M.F., and S.J. have patent "A computer-implemented method, device, system and computer program product for processing anatomic imaging data" pending to University of Zurich related to prior work. 
The remaining authors (A.M., R.Z., L.C., C.J.L.) declare that they have no known competing financial interests or personal relationships that could have appeared to influence the work reported in this paper.

\end{document}